\documentclass[Afour,sageh,times]{sagej}

\usepackage{moreverb,url}

\newcommand\BibTeX{{\rmfamily B\kern-.05em \textsc{i\kern-.025em b}\kern-.08em
T\kern-.1667em\lower.7ex\hbox{E}\kern-.125emX}}

\theoremstyle{definition}
\newtheorem{defn}{Definition}
\newtheorem{problem}{Problem}
\newtheorem{condition}{Condition}

\usepackage{algorithm}
\usepackage{algorithmic}
\usepackage{subfigure}

\def\volumeyear{2015}

\begin{document}

\runninghead{ten Pas, Gualtieri, Saenko, and Platt}

\title{Grasp Pose Detection in Point Clouds}

\author{ten Pas, Andreas\affilnum{1} and Gualtieri, Marcus\affilnum{1} and Saenko, Kate\affilnum{2} and Platt, Robert\affilnum{1}}

\affiliation{\affilnum{1}Northeastern University, Boston, MA 02115\\
\affilnum{2}Boston University, Boston, MA 02215}

\corrauth{Robert Platt, Northeastern University, Boston, MA 02115}

\email{rplatt@ccs.neu.edu}

\begin{abstract}
Recently, a number of grasp detection methods have been proposed that can be used to localize robotic grasp configurations directly from sensor data without estimating object pose. The underlying idea is to treat grasp perception analogously to object detection in computer vision. These methods take as input a noisy and partially occluded RGBD image or point cloud and produce as output pose estimates of viable grasps, without assuming a known CAD model of the object. Although these methods generalize grasp knowledge to new objects well, they have not yet been demonstrated to be reliable enough for wide use. Many grasp detection methods achieve grasp success rates (grasp successes as a fraction of the total number of grasp attempts) between 75\% and 95\% for novel objects presented in isolation or in light clutter. Not only are these success rates too low for practical grasping applications, but the light clutter scenarios that are evaluated often do not reflect the realities of real world grasping. This paper proposes a number of innovations that together result in a significant improvement in grasp detection performance. The specific improvement in performance due to each of our contributions is quantitatively measured either in simulation or on robotic hardware. Ultimately, we report a series of robotic experiments that average a 93\% end-to-end grasp success rate for novel objects presented in dense clutter.
\end{abstract}


\maketitle

\section{Introduction}

Traditionally, robotic grasping is understood in terms of two related 
subproblems: perception and planning. The perceptual component 
estimates the position and orientation (pose) of the object to be 
grasped. The planning component reasons about where/how to move the 
manipulator into a grasp configuration. A typical example of this 
type of approach is the ROS grasp pipeline \citep{chitta_ram2012}. A 
CAD model of the object to be grasped is registered to the point 
cloud or truncated signed distance function. Then, the planning 
component calculates a feasible arm/hand trajectory that grasps the 
localized object. While this type of approach can work well in ideal 
scenarios, it has proven to be surprisingly difficult to apply in 
real world environments. One challenge is that it can be very 
difficult to localize the pose of an object accurately given a noisy 
and partial point cloud~\citep{glover_iros2013}. In addition, the 
approach inherently makes a closed world assumption: that an 
accurate CAD model exists for every object that is to be grasped.

\begin{figure}
\begin{center}
  \includegraphics[height=2.1in]{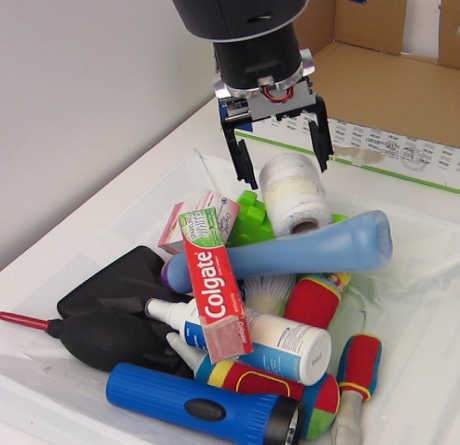}
\end{center}
  \caption{The grasp detection methods proposed in this paper enable us to obtain high grasp success rates in dense clutter such as that shown here.}
  \label{fig:grasp_demo}
\end{figure}

More recently, researchers have proposed various grasp detection 
methods that can be used to localize grasp configurations without 
estimating object pose~\citep{fischinger2012empty,fischinger2013learning,detry2013a,herzog_icra2012,kroemer2012kernel,kappler2015leveraging,tenpas_isrr2015}. These methods take as input a noisy and 
partially occluded RGBD image or point cloud and produce as output 
pose estimates of viable grasps. For the most part, these methods 
treat grasp perception analogously to object detection in computer 
vision: first, they generate a large number of grasp candidates; 
then they evaluate the probability that each candidate is a grasp. A 
classifier or regression system is trained to detect parts of an 
image or a point cloud that can be grasped given large amounts of 
grasp training data. Because these methods detect grasps 
independently of object identity, they typically generalize grasp 
knowledge to new objects well. 

Although grasp detection methods are promising, they have not yet been 
demonstrated to be reliable enough to be used widely. Many grasp 
detection methods achieve grasp success rates (grasp successes as a 
fraction of the total number of grasp attempts) between 75\% and 95\% 
for novel objects presented in isolation or in light 
clutter~\citep{fischinger2012empty,fischinger2013learning,detry2013a,herzog_icra2012,kroemer2012kernel,kappler2015leveraging,tenpas_isrr2015}. Not only are these success rates too low for practical grasping applications, but the light clutter scenarios that are evaluated often do not reflect the realities of real world grasping.

\subsection{Contributions}

This paper describes three algorithmic contributions that make grasp detection more accurate and easier to apply in densely cluttered environments: 

\begin{enumerate}

\item We propose a method for generating grasp hypotheses that, relative to prior methods~\citep{herzog2014learning,kappler2015leveraging}, does not require a precise segmentation of the object to be grasped and can generate hypotheses on any visible surface.

\item We propose a new grasp descriptor that incorporates surface normals and multiple views. Relative to previous methods, this method can improve grasp classification accuracy by approximately 10\%.

\item We propose a method of incorporating prior knowledge about object category that can increase grasp classification accuracy by roughly an additional 2\%.

\end{enumerate}

\noindent
We also provide a systematic evaluation of the new approach in dense 
clutter on real robot hardware:

\begin{enumerate}
  \setcounter{enumi}{3}
\item We introduce a benchmark task that enables the experimenter to systematically evaluate grasp success rates in dense clutter. We use this benchmark to evaluate multiple ablations of our algorithm on a robot in our lab.

\item We introduce a new method of measuring grasp detection performance in terms of recall at a specified high precision, i.e. performance when the number of false positives is constrained to be very small.

\item We introduce and evaluate a method of detecting grasps on a specific object of interest by combining object and grasp detection. 

\end{enumerate}


\vspace{0.1in}


The contributions above incorporate and extend work from two prior conference publications~\citep{tenpas_isrr2015,Gualtieri_iros2016}. In \cite{tenpas_isrr2015}, we proposed the grasp candidate sampling strategy (Item \#1 above) and used the dense clutter benchmark task (Item \#5 above) for the first time. In Gualtieri et al.~\citep{Gualtieri_iros2016}, we proposed the grasp descriptor (Item \#2), the idea of using prior knowledge about object category to improve classification accuracy (Item \#3), and the recall at high precision metric (Item \#5). 

The current paper extends the above work in several ways. First, we describe the algorithms in more detail. Second, we add new results that compare our algorithm with two different ablations: the version of the algorithm where we eliminate the selection strategy and the version where we eliminate the classification step (Table~\ref{table:results}). Third, we add additional results evaluating the improvement in grasp classification accuracy that can result from using category knowledge (Figure~\ref{fig:compareCategoryInfo} (b)). Fourth, we add runtime results that quantify how long it takes to run our algorithm in typical circumstances. Fifth, we propose and evaluate a baseline method of combining object and grasp detection (Item \#6).



\subsection{Comparison to related work}

Grasp detection is distinguished from other approaches to robot grasping because it attempts to detect local grasp surfaces directly from sensor data rather than detecting objects first and then using that information to plan a grasp. This idea originated with Saxena \textit{et al.}, who developed a system for using machine learning to detect grasp points in an image~\cite{saxena_ijrr2008} given a corpus of hand-labeled training data.


\noindent
\textbf{3DOF Grasp Detection:} There are several grasp detection methods that detect grasps in a three-dimensional space, e.g. the space of $x, y, \theta$ configurations in an image. For example, Jiang {\em et al.} models a graspable geometry as an oriented rectangle in an RGBD image~\citep{jiang_icra2011}. In order to perform one of these grasps, the gripper must approach the grasp target from a direction roughly orthogonal to the image. Several other approaches fall into this category as well including~\citep{lenz_rss2013,pinto2015supersizing,redmon2015real,fischinger2012empty}. In particular, Redmon and Angelova use the same dataset as Lenz \textit{et al.} above, but pose grasp detection as a regression problem and solve it using a convolutional neural network (CNN)~\citep{redmon2015real}. Pinto and Gupta's work is notable because their training data comes from on-line experience obtained by the robot during an automated experience-gathering phase rather than from hand-labeled grasps~\citep{pinto2015supersizing}. The work of \cite{fischinger2012empty} goes beyond standard 3DOF approaches by iteratively running a 3DOF grasp detector on a set of different planes that essentially ``view'' the scene from different angles, thereby essentially enabling detection in 6DOF.

\noindent
\textbf{Grasp Templates:} The prior work perhaps most similar to the approach proposed here is the template-based approach of \cite{herzog_icra2012,herzog2014learning}. In that work, the object is roughly segmented from the background and a convex hull is cast around the segmented points. Grasp candidates are generated at the center of each facet in the convex hull at a discrete number of orientations about the approach vector. Each grasp candidate is associated with a set of nearby points that are projected onto a plane and labeled as either object, background, occluded, or void. \cite{kappler2015leveraging} adopt this approach by using a bounding box around the object rather than the convex hull and they allow for translations in addition to rotations on each face (for a total of 16 possible candidate poses per side of the bounding box). Given grasp candidates generated this way, these methods encode each grasp as a single multi-channel image where each pixel denotes the depth along the approach vector and its category, i.e. object, background, occluded, void. While our method contains many similarities to the Herzog/Kappler approaches as described above, there are a few key differences that should be highlighted. First, whereas the Herzog/Kappler method involves segmenting the object from the background and taking a convex hull or a bounding box around the object (see the section on \textit{Grasp Heightmaps} in~\cite{herzog2014learning}), our method does not. Second, whereas those methods only consider grasp candidates located at the center of a facet in the bounding box or convex hull, our method can generate grasp candidates on any visible surface of the object. Third, whereas those methods encode the grasp using a heightmap taken from a single perspective, our method also incorporates surface normals and encodes multiple viewpoints into a single descriptor. The results in this paper quantify the relative advantages of points two and three above.

\noindent \textbf{Grasp representations based on kernel density estimation:} Another set of approaches related to the current work uses kernel representations to encode the local geometry of object surfaces. For example, \cite{detry2013a} developed an approach to grasp detection based on searching for local object surfaces that are similar to one or more members of a set of grasp prototypes. Similarity is measured in terms of the inner product between a kernel density estimator over points on the local object surface and the same estimator for each of the prototypes. Local object surfaces that are similar to the grasp prototypes are deemed likely to be good grasps. \cite{kopicki2014a} use kernel density estimators in a similar way, but they extend the approach to multifingered hands. Finally, \cite{kroemer2012kernel} propose using kernel density estimators to predict the presence of various types of manipulation affordances. An key drawback evident in all of the works cited above~\citep{detry2013a,kopicki2014a,kroemer2012kernel} is that the approach is computationally expensive. In order to find a good grasp at test time, the approach generally requires re-evaluating a similarity metric expressed over the cross product of points in the prototype and the candidates. For example, \cite{detry2013a} reports that it took 18 seconds on average to match a single grasp prototype against candidates on a single object. In general, deep learning based methods (such as that proposed here) that evaluate a neural network once per candidate are much faster (see our runtime results in the experiments section of this paper).

\section{Problem Statement}

Given a point cloud and a description of the geometry of a robotic hand, the grasp pose detection problem is to identify hand 
configurations from which a grasp would be formed if the fingers 
were to close. Let $\mathcal{W} \subseteq \mathbb{R}^3$ denote the 
robot workspace and let $\mathcal{C} \subset \mathcal{W}$ denote a 
set of points in the 3-D point cloud perceived by one or more depth 
sensors registered to the robot workspace. We assume that each point 
in the cloud is paired with at least one viewpoint (camera location) 
from which that point was observed, $\Lambda: \mathcal{C} 
\rightarrow \mathcal{V}$, where $\mathcal{V} \subset \mathbb{R}^3$ 
denotes the set of viewpoints. We will refer to the triple, 
$\mathbb{C} = (\mathcal{C},\mathcal{V},\Lambda)$, as the {\em 
viewpoint cloud}. 

In this paper, we simplify the problem by requiring a two finger hand or a hand that functions like a two finger hand as defined below.

\begin{defn}[{\bf (Two Finger Hand)}]
A {\em two finger hand} is one that can be modelled by two contact surfaces that move toward and away from each other along a one-dimensional manifold.
\label{defn:1}
\end{defn}

\noindent A trivial example of a two finger hand is a typical parallel jaw gripper. Here, contact surfaces on the two grippers are mechanically constrained to move along a single degree of freedom. However, the concept is applicable to a more complex hand such as the Barrett hand where two fingers close in opposition to the third. The two fingers that oppose the third can be modeled as a single contact surface, i.e. a virtual finger~\citep{iberall_book}. Because we only allow two finger hands, the 6-DOF pose of the hand relative to the object is sufficient to calculate the finger contact locations that will result from closing the hand in a given pose relative to an object. As a result, it will be possible to evaluate whether a grasp exists given only the 6-DOF hand pose.

In order to express the grasp pose detection problem, we first need 
to define a grasp. We use the traditional notion of force closure. A 
contact configuration is a {\em force closure} grasp when the 
contact wrenches ({\em i.e.} the forces and torques that can be 
exerted by the contacts) span the origin. For a detailed discussion 
of force closure, see~\cite{murray1994mathematical}. It is assumed 
that the viewpoint cloud encodes partial and noisy information about 
a set of $N$ underlying objects, $\mathcal{O} = O_1, \dots, O_N$. 
The problem of grasp pose detection is to identify force closure grasps in a region of interest, $\mathcal{R} \subset SE(3)$~\footnote{$SE(3) = \mathbb{R}^3 \times SO(3)$ denotes the group of 3D poses.}, of objects in $\mathcal{O}$ by observing only the viewpoint cloud.

\begin{problem}[{\bf (Grasp pose detection)}]
\label{problem:gpd}
Given a viewpoint cloud, $\mathbb{C}$, a region of interest, 
$\mathcal{R} \subset SE(3)$, and a two finger hand, the problem of {\em 
grasp pose detection} is to find one or more 6-DOF hand poses, $h \in 
\mathcal{R}$, such that a force closure grasp will be formed with respect to some 
object, $O \in \mathcal{O}$, when the hand closes.
\end{problem}

\section{Overview of the Grasp Pose Detection Algorithm}

\begin{algorithm}
\caption{Grasp Pose Detection}
\vspace{0.05in}
{\bf Input:} a viewpoint cloud, $\mathbb{C}$; a region of interest, 
$\mathcal{R}$; a hand, $\Theta$; a positive integer, $N$ \\
{\bf Output:} a set of 6-DOF grasp candidates, $H \subset \mathcal{R}$\\
\vspace{-0.15in}
\label{alg:overall}
\begin{algorithmic}[1]
\STATE $\mathbb{C}' = \mbox{PreprocessCloud} (\mathbb{C})$
\STATE $\mathcal{R} = \mbox{GetROI} (\mathbb{C}')$
\STATE $S = \mbox{Sample}(\mathbb{C}',\mathcal{R},\Theta,N)$
\STATE $I = \mbox{Encode}(S,\mathbb{C}',\mathcal{R},\Theta)$
\STATE $H = \mbox{Score}(I)$
\STATE $g = \mbox{SelectGrasp}(S,H)$
\end{algorithmic}
\label{alg:1}
\end{algorithm}

Our algorithm follows the steps shown in Algorithm~\ref{alg:1}. Step 1 preprocesses the viewpoint cloud. Step 2 identifies a region of interest (ROI), $\mathcal{R}$, where the grasp will occur. Step 3 samples $N$ (several thousand) grasp candidates from the ROI where each candidate is a 6-DOF hand pose. Step 4 encodes each grasp candidate as a stacked multi-channel image. Step 5 assigns each candidate a score using a four-layer convolutional neural network that indicates how likely the candidate is to be a grasp. Step 6 selects a grasp for execution based on the score evaluated in Step 5 and other considerations related to the suitability of the grasp.

In the following sections, we discuss Steps 3, 4, 5, and 6 of Algorithm~\ref{alg:1} in detail. We do not discuss Steps 1 and 2 any further. Step 1 compresses and denoises the point cloud by voxelizing, removing outliers, and performing other standard steps. Anything that can be done to reduce noise or errors in the point cloud should be performed in Step 1. Step 2 identifies a region of interest where we are interested in finding a grasp. It is important to note that this does not necessarily mean segmenting the object from the background. The ROI could include a set of multiple objects or all objects in a scene. Or, it could be a rough approximation of the location of the object to be grasped found using an object detection method such as that described at the end of this paper.

\section{Sampling Grasp Candidates}

The goal of Step 3 in Algorithm~\ref{alg:1} is to find a large set of grasp candidates ({\em i.e.} 6-DOF hand poses) where a grasp {\em might} be located. We would like these grasp candidates to be distributed over the graspable portions of the object surface as evenly as possible. We will refer to a grasp candidate, $h \in \mathcal{R}$, as simply a 
``hand''. Let $B(h) \subseteq W$ denote the volume occupied by the 
hand in configuration $h \in \mathcal{R}$, when the fingers are 
fully open. Define the {\em closing region}, $C(h) \subseteq 
\mathcal{W}$, to be the volumetric region swept out by the fingers 
when they close. We sample grasp candidates that satisfy the 
following two conditions:

\begin{condition}
\label{condition:1}
The body of the hand is not in collision with the point cloud: $B(h)
\cap \mathcal{C} = \emptyset$,
\end{condition}

\begin{condition}
\label{condition:2}
The closing region of the hand contains at least one point from the 
cloud: $C(h) \cap \mathcal{C} \neq \emptyset$.
\end{condition}

\noindent
Assuming a noisy point cloud, neither of these conditions is 
strictly necessary or sufficient. However, hands that do not satisfy 
either condition are poor grasp candidates. In particular, while it 
is possible that a grasp might exist that did not satisfy Condition~
\ref{condition:2}, the grasped surfaces would be completely 
invisible to the sensor and it would therefore be difficult to 
accurately predict whether the hand were a grasp.

\begin{algorithm}
\caption{Sample($\mathbb{C}, \mathcal{R}, \Theta, N$)}
\vspace{0.05in}
{\bf Input:} a viewpoint cloud, $\mathbb{C}$; a region of interest, 
$\mathcal{R}$; a gripper, $\Theta$; a grid, $G$; a positive integer, $N$ \\
{\bf Output:} a set of 6-DOF grasp configurations, $H \subset \mathcal{R}$\\
\vspace{-0.15in}
\label{alg:grasphyp}
\begin{algorithmic}[1]
\STATE $H = \emptyset$
\FOR{i = 1 to N}
\STATE Sample $p \in \mathcal{C} \cap \mathcal{R}$ uniformly randomly
\STATE Calculate reference frame, $F(p)$
\FORALL{$(y,\phi) \in G$} 
\STATE $x^* = \min_{x \in \mathbb{R}}$ such that $B(h_{x,y,\phi}(p)) \cap \mathcal{C} = \emptyset$
\IF{$C(h_{\phi,x,y^*}) \cap \mathcal{C} \neq \emptyset$}
\STATE {$H = H \cup h_{\phi,x,y^*}$}
\ENDIF
\ENDFOR
\ENDFOR
\end{algorithmic}
\label{alg:sample}
\end{algorithm}

\begin{figure}
\begin{center}
  \includegraphics[height=1.1in]{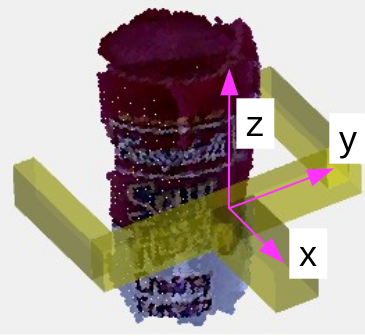}
\end{center}
  \caption{Coordinate frame of $F(p)$ with the hand superimposed at the origin.}
  \label{fig:hand_coordinate_frame}
\end{figure}

Algorithm~\ref{alg:sample} describes the hand sampling process. 
First, we sample $N$ points uniformly at random from $\mathcal{C} 
\cap \mathcal{R}$, the portion of the cloud that intersects the 
region of interest (Steps 2 and 3). Then, for each sampled point, we 
calculate a local reference frame at $p$ by evaluating the 
Eigenvectors of the matrix \[ M(p) = \sum_{q \in \mathcal{C} \cap 
B_r(p)} \hat{n}(q) \hat{n}(q)^T, \] where $\hat{n}(p)$ denotes the outward 
pointing unit surface normal at $p$, calculated using standard methods, 
and $B_r(p)$ denotes the $r$-ball about the point $p$ (Step 4). Let 
$F(p) = \left[ v_3(p) v_2(p) v_1(p) \right]$ denote the orthogonal 
reference frame (known as a Darboux frame) at $p$ where where 
$v_1(p)$ corresponds to the largest Eigenvalue of $M(p)$ and $v_3(p)$
corresponds to the smallest Eigenvalue. The vector $v_3(p)$ is a 
smoothed estimate of the surface normal at $p$, $v_1(p)$ is a 
smoothed estimate of the axis of minor principal curvature, and $v_2(p)$ is a smoothed estimate of the axis of major principal curvature. $F(p)$ is 
rotated by 180 degrees as needed to make $v_3(p)$ point outward from 
the object surface. Figure~\ref{fig:hand_coordinate_frame} shows 
this coordinate frame with the gripper superimposed at the origin.

\begin{figure}
\begin{center}  
  \subfigure[]{\includegraphics[height=0.8in]{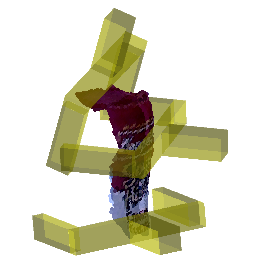}}
  \subfigure[]{\includegraphics[height=0.8in]{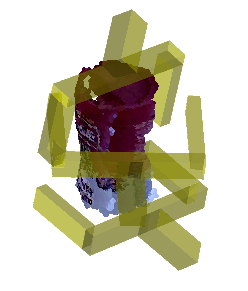}}
  \subfigure[]{\includegraphics[height=0.8in]{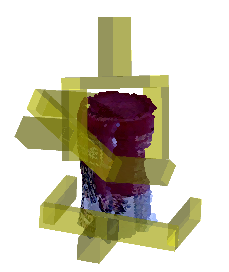}}
  \subfigure[]{\includegraphics[height=0.8in]{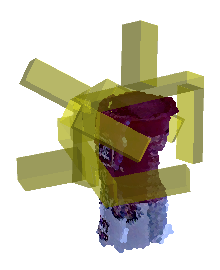}}
\end{center}
  \caption{Illustrations of grasp candidates found using our algorithm. Each image shows three examples of a gripper placed at randomly sampled grasp candidate configurations.}
  \label{fig:sampling}
\end{figure}

Next, in the reference frame of each sample, $p \in \mathcal{C} \cap 
\mathcal{R}$, we perform a local grid search for hands that satisfy 
Conditions 1 and 2 (Steps 5--10). We search a two dimensional grid, 
$G=Y \times \Phi$, where $Y$ and $\Phi$ are discretized sets of 
values along the $y$ axis and about the $z$ axis in the $F(p)$ 
reference frame (in our experiments, $|Y|=10$ and $|\Phi|=8$). For each $(y,\phi) \in G$, we apply the 
corresponding translation and rotation relative to the $F(p)$ 
coordinate frame. Then, we ``push'' the hand forward along the 
negative $x$ axis until one of the fingers or the hand base makes contact with the 
point cloud. Let $T_{x,y,\phi} \in \mathbb{R}^{4 \times 4}$ denote the 
Homogeneous transform that describes a translation of $x,y$ in the 
$x,y$ plane and a rotation of $\phi$ about the $z$ axis. The 
reference frame of the hand at an offset of $x,y,\phi$ relative to 
$F(p)$ is $F(p) T_{x,y,\phi}$. For each $(y,\phi) \in G$ , let 
$F(h_{y,\phi})$ be the hand at gridcell $y,\phi$ that has been 
pushed forward until making contact with the cloud: $F(h_{y,\phi}) = 
F(p) T_{x^*,y,\phi}$, where $x^* = \min_{x \in \mathbb{R}}$ such 
that $B(h_{x,y,\phi}(p)) \cap \mathcal{C} = \emptyset$  (Step 6). A 
hand, $h_{y,\phi}$, is added to the set of grasp candidates, $H$, if 
the closing region, $C(h_{y,\phi})$ contains at least one point in 
the cloud (Steps 7-9). Figure~\ref {fig:sampling} shows examples of 
grasp candidates found using this method for a partially occluded 
point cloud of a soup can.


\subsection{Comparison With Other Grasp Sampling Methods From The Literature}

Most grasp detection algorithms use some method for generating grasp 
proposals. A number of methods generate proposals exhaustively using 
a sliding window~\citep 
{saxena_ijrr2008,jiang_icra2011,fischinger2012empty,fischinger2013learning}. \cite{pinto2015supersizing} cluster foreground pixels using 
background subtraction on the RGBD image. The sampling method from 
the literature most similar to ours method is perhaps that of~\cite 
{herzog_icra2012} (later adopted by \cite{kappler2015leveraging}). 
That method works by segmenting the object to be grasped from the 
rest of the point cloud and calculating the convex hull. For each 
face in the convex hull, a coordinate frame is created at the center 
of each face and oriented such that the $x$ axis points outward 
(similar to $F(p)$ in our method as shown in Figure~\ref 
{fig:hand_coordinate_frame}). A one-dimensional grid search is 
performed over a set of angular displacements about the $x$ axis and 
the hand is ``pushed'' toward the object (similar to what we do in 
Step 6 of Algorithm~\ref {alg:sample}). 


There are two main differences between the method described above and ours. First, whereas \cite{herzog_icra2012} segment the object, cast a convex hull around it, and only sample grasps at the center of each face in the hull, we sample grasps on all visible surfaces. This makes our sampling method more practical in cluttered scenarios where object segmentation can be challenging. Second, whereas \cite{herzog_icra2012} and \cite{kappler2015leveraging} perform a one-dimensional grid search over hand orientations about the $x$ axis in the $F(p)$ coordinate frame, we perform a two-dimensional grid search over positions along the $y$ axis and orientations about the $z$ axis in $F(p)$.

\begin{figure}[t]
\begin{center}  
  \subfigure[]{\includegraphics[height=0.8in]{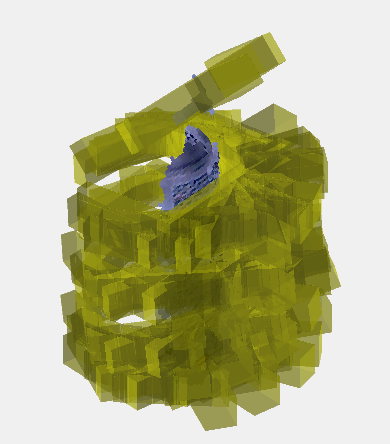}}
  \subfigure[]{\includegraphics[height=0.8in]{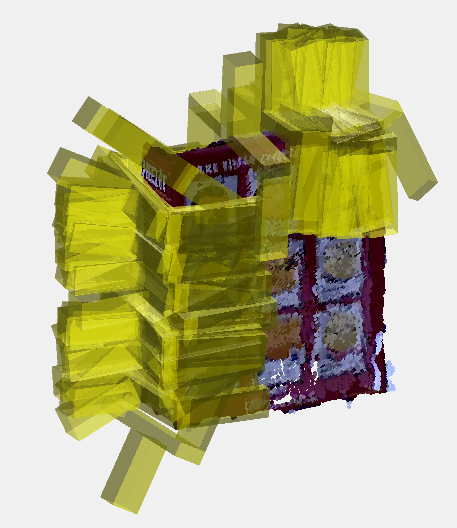}}
  \subfigure[]{\includegraphics[height=0.8in]{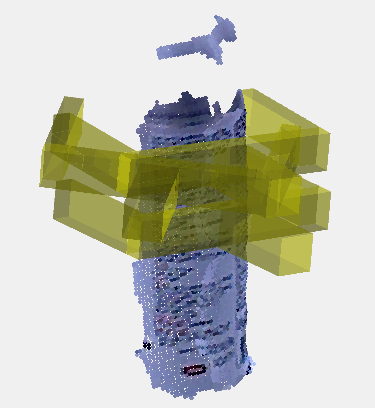}}
  \subfigure[]{\includegraphics[height=0.8in]{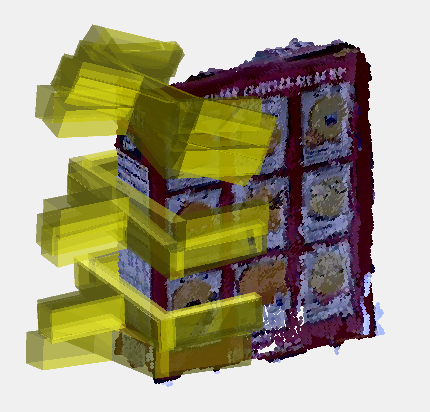}}
\end{center}
  \caption{Force closure grasps sampled using our method where $\Phi$ is interpreted as an orientation about the $z$ axis of the $F(p)$ coordinate frame ((a) and (b)) and a variation of our method where $\Phi$ is interpreted as an orientation about the $x$ axis (as in \cite{herzog_icra2012}) ((c) and (d)). Our method generates a larger number of force closure grasps spread more uniformly over the graspable parts of the object surface.}
\label{fig:sampling_comparison}
\end{figure}

The discussion above begs the question whether it is better to search over orientations about the $z$ axis or the $x$ axis. To answer this question, we compared our method where we search over a grid, $G = Y \times \Phi$ ($|Y|=80$, $|\Phi|=8$), with $\Phi$ interpreted as an orientation about the $z$ axis with a variant of the method where $\Phi$ was interpreted as an orientation about the $x$ axis. We compared the methods for two objects: a box-like object (the Cheezits from BigBird~\cite{}) and a cylindrical object (the Krylon Crystal Clear from BigBird), each viewed from 20 different perspectives. We found that on average, our method produced at least three times the number of true positive grasp candidates (122 versus 22 positives for the Cheezeits and 259 versus 77 for the Krylon). Figure~\ref{fig:sampling_comparison} illustrates representative results. The true positive grasp candidates shown in Figure~\ref{fig:sampling_comparison}(c,d) were found using the method where $\Phi$ was interpreted as an orientation about the $x$ axis as in \cite{herzog_icra2012} whereas those shown in Figure~\ref{fig:sampling_comparison}(a,b) were found using our method. Notice that our method finds more true positive grasp candidates and that they are better distributed over the graspable object surface.

We hypothesize that this improvement in performance is related primarily to the fact that we used a two-fingered hand in this evaluation. Our method samples most densely candidates where the closing region of the hand is orthogonal to the axis of minor principal curvature of the local object surface. For a two-fingered hand, these are the configurations where the object is most likely to fit between the fingers. For example, by orienting the hand orthogonal to the axis of the cylinder in Figure~\ref{fig:sampling_comparison} (a), our method is most likely to generate candidates that fit the cylinder between the fingers. While we expect this sampling strategy to work well for different types of two-fingered hands, these advantages may disappear for hands with more complex closing regions.



\section{Classifying Grasp Candidates}

We frame grasp detection as a binary classification task and solve it using a four-layer convolutional neural network (a CNN). For a given grasp candidate, the input to the CNN is a representation of the observed and occluded object surfaces as seen by a depth sensor in the neighborhood of the candidate (Step 4 of Algorithm~\ref{alg:1}). The output is a prediction of whether or not the candidate is a grasp (Step 5 of Algorithm~\ref{alg:1}). We use the same CNN structure used by LeNet~ \citep{lecun1998gradient}: two convolutional/pooling layers followed by one inner product layer with a rectified linear unit at the output and one more inner product layer with a softmax on the output. The outputs, kernel size, pooling strides, {\em etc.} are all identical with those used by the LeNet solver provided in Caffe~\citep{Jia2014}  (this same structure is used by~\cite {kappler2015leveraging}). Our network was trained using stochastic gradient descent with a learning rate of 0.00025. This section describes the method for representing the grasps to the CNN and our method for creating a dataset with which to train the network.


\subsection{Grasp Candidate Representation}
\label{sect:representation}

\begin{figure}[b]
\begin{center}  
  \subfigure[]{\includegraphics[height=1.2in]{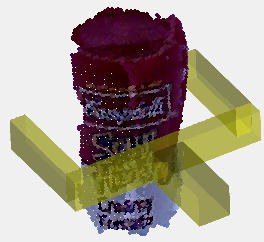}}
  \subfigure[]{\includegraphics[height=1.2in]{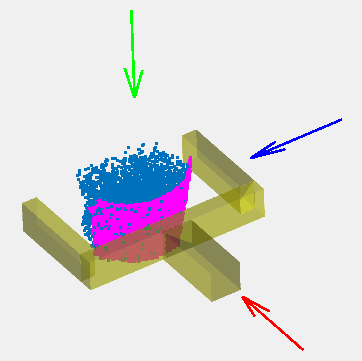}} 
  \\
  \subfigure[]{\includegraphics[height=0.8in]{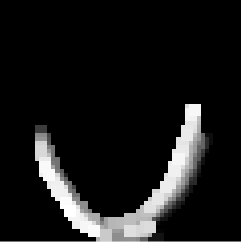}}
  \hspace{0.3in}
  \subfigure[]{\includegraphics[height=0.8in]{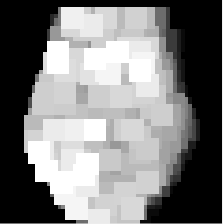}}
  \hspace{0.3in}
  \subfigure[]{\includegraphics[height=0.8in]{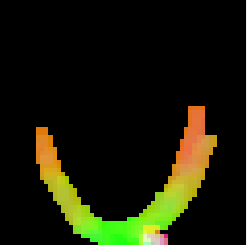}}
\end{center}
  \caption{Grasp representation. (a) A grasp candidate generated from partial point cloud data. (b) Local voxel grid frame. (c-e) Examples of grasp images used as input to the classifier.}
  \label{fig:representation1}
\end{figure}

We represent a grasp candidate to the classifier in terms of the geometry of the observed surfaces and unobserved volumes contained within the closing region of the gripper, $C(h) \subset \mathcal{W}$. This volume can be arbitrary, but it should be represented in the reference frame of the hand. Our representation is illustrated in Figure~\ref 
{fig:representation1}. Figure~\ref{fig:representation1}(a) shows a 
grasp candidate generated with respect to partial point cloud data 
(from the BigBird dataset~\citep{singh2014bigbird}). Figure~\ref{fig:representation1}(b) shows two sets of points in $R$. One set of points, shown in magenta, shows points in the cloud contained within $R$. The other set of points, shown in blue, are sampled from the portion of $R$ that is unobserved, {\em i.e.}, that is occluded from view by every sensor. Currently, we ignore RGB information associated with the points in the cloud. Notice that this representation depends both on the geometry of the object surface and on the perspective from which it is viewed.

Assume the closing region $C(h)$ is scaled to fit inside the unit 
cube and the points contained within it are voxelized into a $M 
\times M \times M$ grid. For every triple, $(x,y,z) \in [1,M] \times 
[1,M] \times [1,M]$, $V(x,y,z) \in \{0,1\}$ denotes whether the 
corresponding voxel is occupied and $U(x,y,z) \in \{0,1\}$ denotes 
whether the corresponding voxel has been observed. We will further 
assume that each occupied voxel in $C(h)$ is associated with a unit, 
outward-pointing surface normal vector, $\hat{n}(x,y,z) \in S^2$ (
$S^2$ denotes the surface of the 2-sphere), that denotes the 
orientation of the object surface at that point. All of the above 
information can be calculated either from the point cloud with 
associated viewpoints or from a truncated signed distance function.



We want to represent the 3D geometry of the object surface contained 
within the closing region. Inspired by~\cite{su2015multi}, we encode 
this volume using a multiple view representation. Specifially, for a 
hand, $h \in H$, we project the voxels onto planes orthogonal to the 
axes of the hand reference frame, $F(h)$, and pass these to the CNN 
as input. Figure~\ref{fig:representation1}(b) shows the three 
directions from which we view the closing region. The red arrow 
points along the negative $x$ axis of $F(h)$. The blue arrow points 
along the negative $y$ axis. The green arrow points along the 
negative $z$ axis. For each of these three projections, we will 
calculate three images: an averaged heightmap of the occupied 
points, $I_o$, an averaged heightmap of the unobserved region, $I_u$, 
and averaged surface normals, $I_n$. For example, to project onto 
the plane constituted by the hand and major curvature axes, {\em i.e.}, 
the (x,y) plane, these maps are calculated as follows: \[ I_o(x,y) = 
\frac{\sum_{z \in [1,M]} z V(x,y,z)}{\sum_{z \in [1,M]} V(x,y,z)} \] 
\[ I_u(x,y) = \frac{\sum_{z \in [1,M]} z U(x,y,z)}{\sum_{z \in [1,M]}
U(x,y,z)} \] \[ I_n(x,y) = \frac{ \sum_{z \in [1,M]} 
\hat{n}(x,y,z) V(x,y,z) }{\sum_{z \in [1,M]} V(x,y,z)} \] The first 
two images, $I_o$ and $I_u$, are $M \times M$ images (Figure~\ref{fig:representation1} (c,d)). The last 
image, $I_n(x,y)$, is a $M \times M \times 3$ image where the three 
dimensions of the normal vector are interpreted as three channels in 
the image (Figure~\ref{fig:representation1} (e)). Constant scaling and offsets are applied to each of these images to ensure that all pixel values lie between 0 and 255. In the case of $I_n$, we scale the three channels by the same factor so that the $[1,m]$ interval can encode all possible unit normal vectors. All together, we have five channels of information for each of the three projections, for a total of 15 channels.

\section{A Grasp Dataset for Training the CNN}

In order to train the grasp detection CNN, a dataset is needed that pairs grasp candidates with ground truth labels that denote whether or not the candidate is a force closure grasp. Since the representation discussed above depends on both the geometry of the grasp surface and on the perspective from which it was viewed, the dataset should span both of these variables -- it should contain examples of different graspable geometries as seen from different viewpoints. For this paper, we created a dataset of 1.5M exemplars with equal numbers of positive and negative examples comprised of up to 50k labeled grasp exemplars for each of 55 objects in the BigBird dataset~\citep{singh2014bigbird} (out of a total of 125 objects in BigBird). The 55 objects included 29 box-type objects, 16 cylindrical-type objects, and 10 other objects for which: 1) a complete mesh exists for the object in the dataset; 2) the object can be grasped by a parallel jaw gripper that can open by at most 10cm. It should be noted that although we have created a dataset with 1.5M exemplars, we only use about 300k of this data (randomly sampled from the whole) in any single training/test session. This is because it was a computational burden to load the entire dataset in our current setup and we simply did not need all the data to train our four-layer CNN.

\begin{figure}[h]
\begin{center}  
    \includegraphics[height=2.0in]{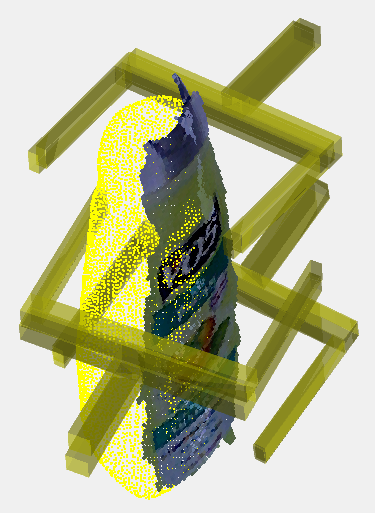}
\end{center}
  \caption{During creation of the dataset, grasp candidates are sampled from point clouds. Then, a mesh registered to the point cloud is used to evaluate whether the candidate is force closure.}
  \label{fig:cloudmesh}
\end{figure}

To create this dataset, it is necessary to obtain complete mesh models of graspable objects paired with point clouds created by viewing each object from different perspectives. One approach is to start with object CAD models and to simulate the point cloud that would be observed from different perspectives. However, as we will show in the next section, training data obtained from simulated point clouds does not transfer to point clouds obtained using real sensors as well as one might expect. As a result, we rely heavily upon the BigBird dataset because it pairs reconstructed object meshes with real point clouds produced by an Asus Xtion Pro depth sensor taken from a large number of different perspectives. For each of the 55 objects, we generated 20 registered clouds created by registering together clouds taken from two different viewpoints 53 degrees apart. This relative configuration of the two point clouds reflects the sensor configuration on our robot (Figure~\ref{fig:baxter_pic}(a)). See Figure~\ref{fig:cloudmesh} for an example of a point cloud registered with the mesh. For each registered cloud, we sampled a large number of grasp candidates and evaluated whether each was a force closure grasp (see below). Data from all 20 views were concatentated together, and a sufficient number of non-grasps were discarded so that the number of positive and negative exemplars for each object was balanced. Ultimately, we generated up to 50k labeled grasp exemplars for each object. While we use BigBird in this paper, any object dataset that pairs RGBD images with mesh models of the object geometry will work.


\subsection{Labeling grasp candidates}

In order to create the dataset, we need a way to label each grasp candidate as a good grasp or not. Recall that each grasp candidate is associated with a 6-DOF hand pose. And, recall that we restrict ourselves to the set of all two finger hands as described in Definition~\ref{defn:1}. As a result, it is possible to label each candidate as a good grasp or not by evaluating whether a force closure grasp~\citep{murray1994mathematical} would be formed if the fingers were to close from the given 6-DOF hand pose. Actually, there is nothing about our method that specifically requires us to use force closure \textit{per se}. As some have noted recently~\citep{roa2015grasp,kappler2015leveraging}, the force closure condition may not always be a good predictor of whether a given grasp will succeed in practice. Our only requirement is that some method exists for checking whether a good grasp would be formed from a given grasp candidate pose, given a full mesh model of the object. Nevertheless, in the experiments reported in this paper that use a two finger parallel jaw gripper, the standard force closure analysis seems to work well.

Since we have constrained consideration to two finger hands that use only two contact surfaces, we opt for a special case of force closure known as a frictionless antipodal grasp. A set of contacts is a \textit{frictionless antipodal grasp} when the applied frictionless contact wrenches would balance each other out if they were each to apply unit forces~\citep{platt_tro10,murray1994mathematical}. This is a more conservative condition than force closure in the sense that any frictionless antipodal grasp is also a force closure grasp for any non-zero coefficient of coulomb friction (assuming soft contacts). In addition, it is robust to small contact perturbations. For any positive coefficient of friction and for any object with a Lipschitz continuous surface, there exists a neighborhood of contact configurations around a frictionless antipodal grasp that are also force closure.


One challenge in evaluating whether a hand configuration is a frictionless antipodal grasp is that the available object meshes can be noisy, e.g. the object meshes in the BigBird dataset are noisy because they are reconstructed from actual sensor data. We address this by assuming that each vertex in the mesh is subject to a small amount of position error (1mm in our experiments). We evaluate whether there exists any perturbation of the vertices within the 1mm bound for which an equilibrium grasp would exist. We chose 1mm for this position constant in our experiments because it was qualitatively the smallest constant that seemed to smooth out the unit frictionless equilibrium condition as a function of contact positions. For the case of a parallel jaw gripper (as in our experiments), this reduces to identifying small contact regions in each finger where contact might be established and evaluating whether the frictionless antipodal condition described above holds for any pair of contacts in these regions.

\section{Improving Classification Accuracy}

We have found that a number of design decisions can have a significant effect on grasp detection accuracy including: the grasp representation used, whether the CNN is pretrained using data derived from CAD models, and whether information about the identity of the object being grasped is used. We preface this analysis with a discussion of how grasp detection performance should be evaluated.

\subsection{Measuring recall-at-high-precision for grasp pose detection}

\begin{figure}[b]
\begin{center}  
  \subfigure[]{\includegraphics[height=1.5in]{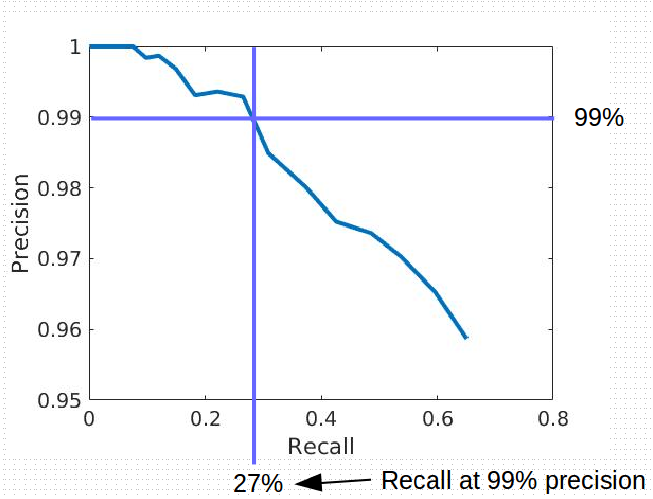}}
  \subfigure[]{\includegraphics[height=1.5in]{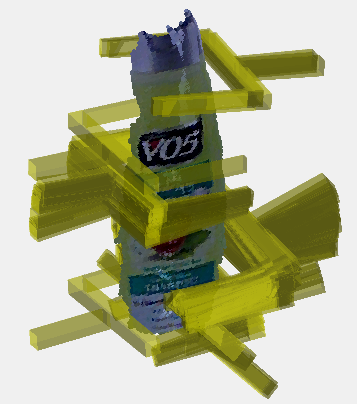}} 
\end{center}
  \caption{Example of recalling grasps at high precision. (a) Precision-recall curve. (b) Grasps recalled at 99\% precision.}
  \label{fig:recall_at_precision}
\end{figure}

We propose  a  new  method  of measuring grasp detection performance in terms of recall at a specified high precision. Typically, classification performance is measured in terms of accuracy -- the proportion of predictions made by the classifier that match ground truth. Most grasp pose detection systems described in the literature achieve something between 75\% and 95\% grasp classification accuracy~\cite{fischinger2012empty,fischinger2013learning,detry2013a,herzog_icra2012,kroemer2012kernel,kappler2015leveraging,tenpas_isrr2015}. Unfortunately, this accuracy number alone does not give us a good indication of whether the resulting grasp pose detection system will have a high grasp success rate. The key question is whether a particular grasp pose detection system can detect grasps with high precision. Precision is the proportion of all positives found by the classifier that are true positives. In grasp pose detection, the cost of a false positive is high because it can cause a grasp attempt to fail. As a result, we want to travel along the precision-recall curve and reach a point with very high precision ({\em i.e.} very few false positives). This amounts to adjusting the classifier acceptance threshold. Setting the threshold very high will result in a high precision, but it will reduce recall -- the proportion of all true positives found by the classifier. Therefore, a key metric for grasp pose detection is recall-at-high-precision. If the number of false positives is constrained to be less than a desired threshold percentage (say 1\% fase discovery rate, i.e. 99\% precision), what proportion of all true grasps will be detected, i.e. what is the recall? This is illustrated in Figure~\ref{fig:recall_at_precision}(a). For a particular shampoo bottle instance, we can recall 27\% of the grasps at 99\% precision. The key insight is that since grasp pose detection systems can detect hundreds of grasps for a single object, we do not need to recall {\em all} of the grasps in order to have lots of choices about which grasp to execute. This is illustrated in Figure~\ref {fig:recall_at_precision}(b). Although we are only detecting 27\% of all true positives, there are still plenty of alternatives.

\subsection{Comparison Between Different Representations}
\label{sect:comparison}

\begin{figure}[h]
\begin{center}  
    \includegraphics[height=2.5in,trim={0.4in 0.1in 0.6in 0.5in},clip]{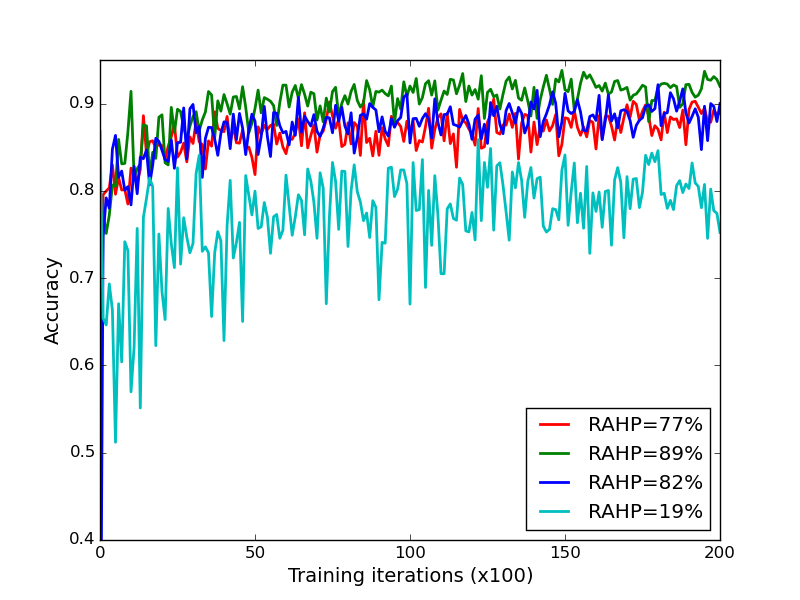}
\end{center}
  \caption{Classification accuracy obtained using different grasp candidate representations. Green: combined 15-channel representation. Blue: same as green but without the occlusion channels. Red: the representation used in our prior work~\cite{tenpas_isrr2015}. Cyan: the representation used in both Kappler {\em et al.}~\cite{kappler2015leveraging} and Herzog {\em et al.}~\cite{herzog_icra2012}. The legend shows the recall-at-high-precision (RAHP) metric for each of these representations for 99\% precision.}
  \label{fig:comparisonOfRepresentations}
\end{figure}

Grasp classification accuracy depends upon how the candidates are represented to the convolutional neural network (CNN) that does the classification. We compare the following four representations:

\vspace{0.25cm}


\noindent
\textbf{Representation \#1:} The 15-channel representation described in this paper (green line in Figure~\ref{fig:comparisonOfRepresentations}).

\vspace{0.25cm}

\noindent
\textbf{Representation \#2:} A 12-channel ablation of the 15-channel representation where we delete the three unobserved region channels, i.e. delete one $I_u$ channel for each of the three views (blue line in Figure~\ref{fig:comparisonOfRepresentations}).

\vspace{0.25cm}

\noindent
\textbf{Representation \#3:} The 3-channel representation used in our prior work~\citep{tenpas_isrr2015} (red line in Figure~\ref{fig:comparisonOfRepresentations}). This representation consists only of $I_n$ projected onto the plane orthogonal to the negative $z$ axis of $F(h)$ (the plane orthogonal to the green arrow in Figure~\ref{fig:representation1} (b)). Whereas \cite{tenpas_isrr2015} encodes this 3-channel image using HOG features, we supply the image directly to a similar CNN (convolutional neural network) as that used with the 15-channel representation.

\vspace{0.25cm}

\noindent
\textbf{Representation \#4:} The 3-channel representation used in both \cite{kappler2015leveraging} and \cite{herzog_icra2012} (cyan line in Figure~\ref{fig:comparisonOfRepresentations}). That representation is comprised of three channels of information projected along the hand approach axis. One channel is $I_o$. The second channel is $I_u$. The third channel describes the unoccupied voxels in the space: $I_f = \overline{I_u \cup I_o}$. 


\vspace{0.25cm}

To do the comparison, we created a dataset of grasp candidates equally balanced between positives and negatives by randomly sampling approximately 4k grasp candidates from each of 55 BigBird objects. This resulted in a total of 216k candidates with a 185k/31k train/test split over views (for each object, the test set does not contain any exemplars derived from a view that is present in the training set). For each of the four representations, we re-train the convolutional neural network starting with random weights and compare classification accuracy on the 31k test set. 

The accuracy of our full 15-channel representation (Representation \#1) as a function of training iteration (in 100s) is shown in green in Figure~\ref {fig:comparisonOfRepresentations}. Since this train/test split is over views, this result describes the accuracy that would be achieved if we knew we would be given one of the 55 BigBird objects, but did not know in advance which one we would get.

The accuracy of the 12-channel representation (Representation \#2) is shown in blue in Figure~\ref{fig:comparisonOfRepresentations}. Notice that we gain approximately an additional 2\% accuracy by using the occlusion information. The recall-at-high-precision (99\% precision) for the 15-channel and the 12-channel representation is 89\% and 82\%, respectively. This comparison is important because the occlusion information is computationally expensive to calculate -- it can double the amount of time taken to encode candidates for classification. This additional computation time must be balanced against the 2\% additional accuracy that it yields. 

The accuracy of the 3-channel representation used in our prior work~\cite{tenpas_isrr2015} (Representation \#3) is shown in red in Figure~\ref{fig:comparisonOfRepresentations}. This representation contains only the surface normals image, $I_n$, for the projection onto the plane orthogonal to the negative $z$ axis of $F(h)$ (the plane orthogonal to the green arrow in Figure~\ref{fig:representation1} (b)). Notice that this representation performs just about as well as the 12-channel, even though it only contains three channels of information. Compared with the 15-channel representation, this representation is much less expensive to compute -- four times faster (see the comparison in Table~\ref{table:runtime}). This suggests that it could be appropriate for time-sensitive applications where it is important to do grasp detection quickly.

Finally, the cyan line in Figure~\ref{fig:comparisonOfRepresentations} (Representation \#4) shows the accuracy obtained using the 3-channel representation used in both \cite{kappler2015leveraging} and\cite{herzog_icra2012}. Although we are using the Herzog representation, we are evaluating on grasp candidates generated using our own method (the same set of candidates used to evaluate the other representations described above). Notice that, on average, this representation obtains at least 10\% lower accuracy than the other representations and only a 19\% recall-at-high-precision. This loss in performance could be caused by the fact that this representation projects onto the plane orthogonal to the surface normal (the plane orthogonal to the red arrow in Figure~\ref{fig:representation1} (b)) rather than the projection from representation \#3. It could also be caused by the fact that this representation does not encode object surface normals. 

We draw a few conclusions from the results described above. First, if the primary objective is to maximize grasp detection accuracy, then the 15-channel representation (Representation \#1) should be used. While it can take longer to calculate the 15-channel descriptor relative to the other representations, this one gets the best accuracy. Second, for time-sensitive applications where it is important to detect grasps quickly, the 3-channel representation (Representation \#3) should be used. This one does almost as well as the 12-channel representation and it can be computed approximately four times faster than the 15 channel representation (see the results on algorithm runtime in the dense clutter experiments section of this paper). Finally, this method should be preferred relative to the 3-channel representation from \cite{kappler2015leveraging} and \cite{herzog_icra2012} because it obtains much better grasp classification accuracy.

\subsection{Pretraining on simulated data}

\begin{figure}
\begin{center}
  \includegraphics[height=2.4in,trim={0.3in 0.1in 0.6in 0.5in},clip]{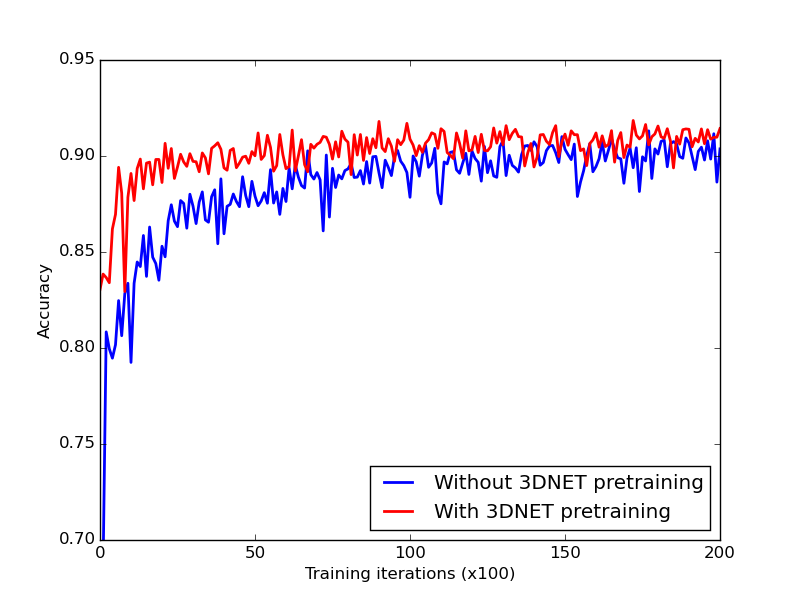}
\end{center}
  \caption{Accuracy with (red) and without (blue) 3DNET pretraining.}
  \label{fig:3dnetpretrain}
\end{figure}

One way to improve classifier accuracy and precision is to create training data using point clouds or truncated signed distance functions created by simulating what a sensor would observe looking at a CAD model. Compared with the amount of real sensor data that is available, there is a huge number of CAD models available online (e.g., 3DNET makes available thousands of CAD models from 200 object categories~(\cite{Wohlkinger2012}). Ideally, we would train using this simulated data. Unfortunately, there are subtle differences between depth images obtained from real sensors and those obtained in simulation that hurt performance. For example, recall the 31K test set derived from BigBird data described in the last section. Our best representation obtained approximately 90\% accuracy over all 55 objects in the BigBird dataset. However, when we train our system for 30000 iterations on 207K exemplars created using 400 object CAD models taken from 16 categories in 3DNET, we obtain only 83\% accuracy on the same test set. While it is possible that different methods of simulating depth images could improve performance, it is likely that a small difference will persist.

One approach to this problem is to pretrain the CNN learning system using simulated data, but to ``finetune'' it on real data more representative of the problem domain at hand. We evaluated this approach by testing on the 216K BigBird dataset described in the last section. We compare the learning curve obtained using the 15-channel representation starting with random network weights with the learning curve obtained using the 3DNET weights as a prior. Figure~\ref{fig:3dnetpretrain} shows the results. The pretrained weights have a strong effect initially: the pretrained network obtains the same accuracy at 4000 iterations as the non-pretrained network obtains after 20000 iterations. However, the importance of the contribution diminishes over time.

\subsection{Using prior knowledge about the object}

Another way to improve grasp detection accuracy is to incorporate prior knowledge of the object to be grasped. One way to do this is to adjust the contents of the training dataset. If we have no prior knowledge of the object to be grasped, then the training dataset might include data from a large and diverse set of objects. If we know the category of the object to be grasped (for example, if we know the object is box-like), then we might train the grasp detector using training data from only box-like objects. Finally, if we know the exact object geometry, then we might use training data derived only from that particular object. In general, one would expect that the more prior knowledge that is encoded into the network this way, the better our classification accuracy will be. We envision obtaining information about the object by using standard object detection methods as proposed later in this paper. Object detection is quickly becoming a reliable technology. Even though it is difficult to detect the precise \textit{pose} of an object to be grasped, a variety of methods exist for locating an object by instance or category label within an image~\cite{Jia2014,girshick2014rcnn}.

\begin{figure}
\begin{center}  
  \subfigure[]{\includegraphics[width=3.5in]{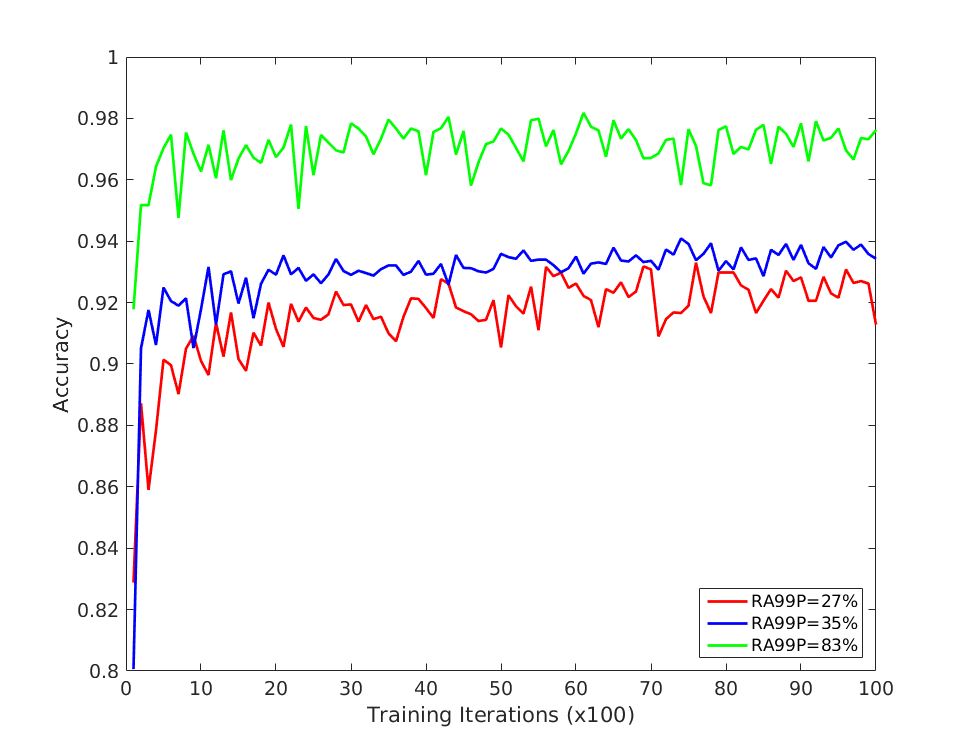}}
  \subfigure[]{\includegraphics[width=3.5in]{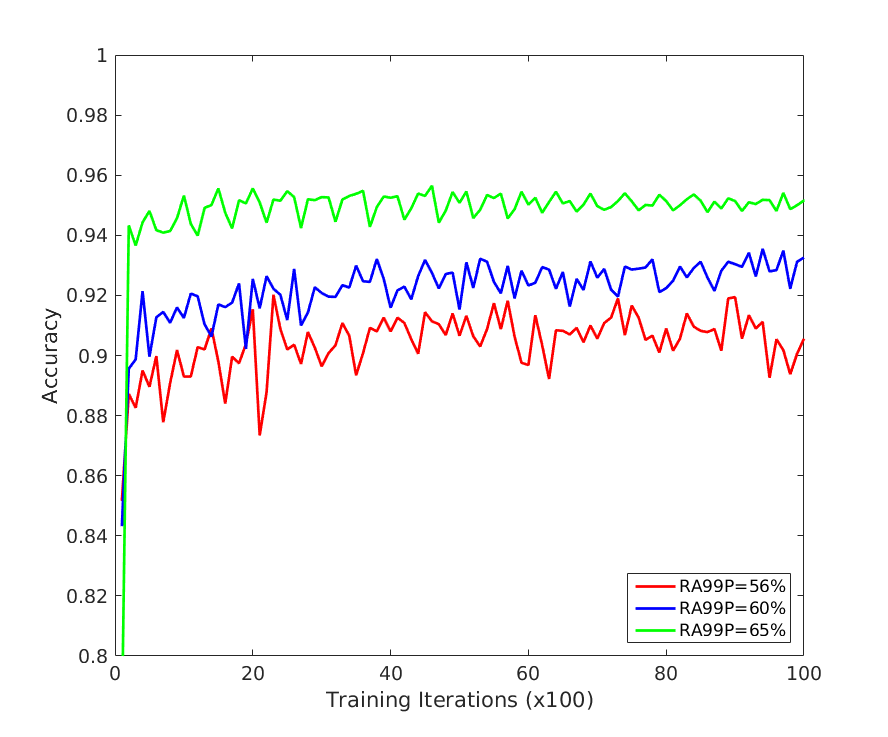}}
\end{center}
  \caption{Grasp detection accuracy given no prior knowledge of the objects (red); given a category knowledge (blue); given the precise geometry of the object (green). (a) shows the comparison for a set of 16 cylindrical objects. (b) shows the same comparison for a set of 29 box-like objects. The legend shows the recall-at-99\%-precision (RA99P) metric for each of these representations.}
  \label{fig:compareCategoryInfo}
\end{figure}

To evaluate the effect of prior information as described above, we performed experiments with two classes of objects included in BigBird: 16 cylindrical-like objects and 29 box-like objects. For each of these two classes, we compared classification accuracy in these three different scenarios. In all cases, we started with a network that was pretrained on 3DNET data. First, we trained a network using training data derived only from the single object in question using a train/test split (45K training and 5K test) on view angle (no one view shared between test and training). Averaged over the 16 cylindrical-like objects, we obtained roughly 97\% classification accuracy and 83\% recall-at-99\%-precision. Averaged over the 29 box-like objects, we obtained 95\% classification accuracy and 65\% recall-at-99\%-precision (the green lines in Figures~\ref{fig:compareCategoryInfo}(a) and (b)). Second, for each of the 16 cylindrical-like objects, we trained the network using data derived from the other 15 objects (leave-one-object-out). This gave us 150K exemplars for training and 10K for test for each cylindrical-like object. Here we obtained approximately 93.5\% accuracy and 35\% recall-at-99\%-precision. Doing the same experiment for the 29 box-like objects (280k training, 10k test for box-like objects), we obtained roughly 93\% accuracy and 60\% recall-at-99\%-precision (the blue lines in Figures~\ref{fig:compareCategoryInfo}(a) and (b)). Finally, for each object, we trained the network using all other objects in the dataset. This gave us 285K training examples and 5K test for each object. For cylinder-like objects, we obtained approximately 92\% accuracy and 27\% recall-at-99\%-precision. For box-like objects, we obtained approximately 90\% accuracy and 56\% recall-at-99\%-precision (the red lines in Figures~\ref{fig:compareCategoryInfo}(a) and (b)). The fact that we achieve such high detection accuracy in the case where we have prior knowledge of the object {\em instance} to be grasped is important because this case has been studied extensively in the literature. The standard approach is to fit a CAD model of the object instance to the point cloud and to register planned grasp from the CAD model to the cloud. However, it is well known that it can be hard to get the standard approach to work reliably in cluttered grasp environments. In comparison, instance-level grasping is extremely accurate (the green lines in Figures~\ref{fig:compareCategoryInfo}(a) and (b)). With 83\% recall-at-99\%-precision, we will detect most good grasps that get sampled. Moreover, a variety of object detection methods are available to locate specific object instances of interest.



\section{Dense Clutter Grasping Experiments}

While it is clear that grasp pose detection can detect grasps accurately, it is important to evaluate the level of grasp performance that can be achieved on an physical robot. Our main evaluation mechanism is the dense clutter benchmark, described below. As is shown in the next section, the version of the system that uses the 3-channel descriptor (Representation \#3 from the section entitled \textit{Comparison Between Different Representations}) runs significantly faster than the one using 15-channel descriptor with only a small resulting deficit in accuracy. As a result, we standardized all the following hardware tests on the 3-channel version.



\subsection{Dense Clutter Benchmark Task}
\label{sec:exp_protocol}

\begin{table}[h]
\small\sf\centering
\caption{Dense clutter benchmark task.\label{table:experimentProtocol}}
\begin{tabular}{l|l}
\toprule
1. & Randomly select 10 objects from the object set.\\
 \hline
2. & Place objects into a box.\\
 \hline
3. & Shake box until sufficiently mixed.\\
 \hline
4. & Pour box contents into tray in front of robot.\\
 \hline
5. & Run grasp algorithm. \\
 \hline
6. & Terminate once any of these events occur:\\
 & i) No objects remain in tray.\\
 & ii) No grasp hypotheses were found after 3 attempts.\\
 & iii) The same failure occurs on the same object 3 times.\\
\bottomrule
\end{tabular}\\[10pt]
\end{table}

\begin{figure}[h]
\begin{center}
  \subfigure[]{\includegraphics[height=2in]{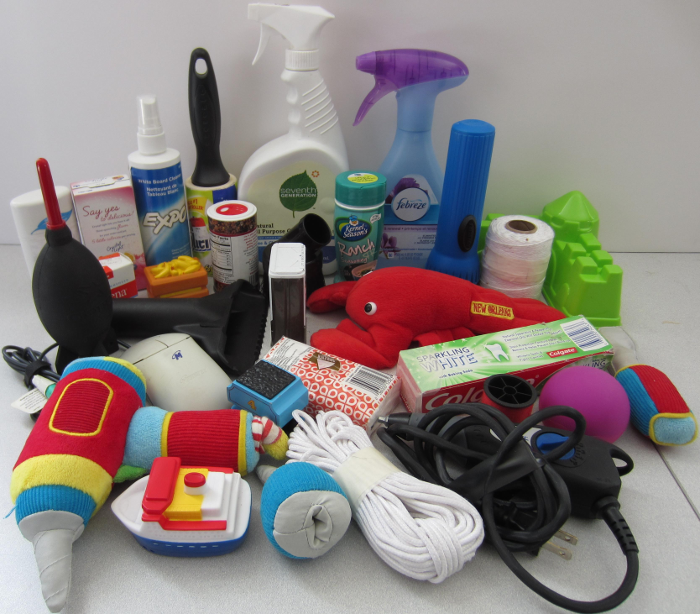}}
  \subfigure[]{\includegraphics[height=1.6in,trim={0.2in 0in 0in 1.0in},clip]{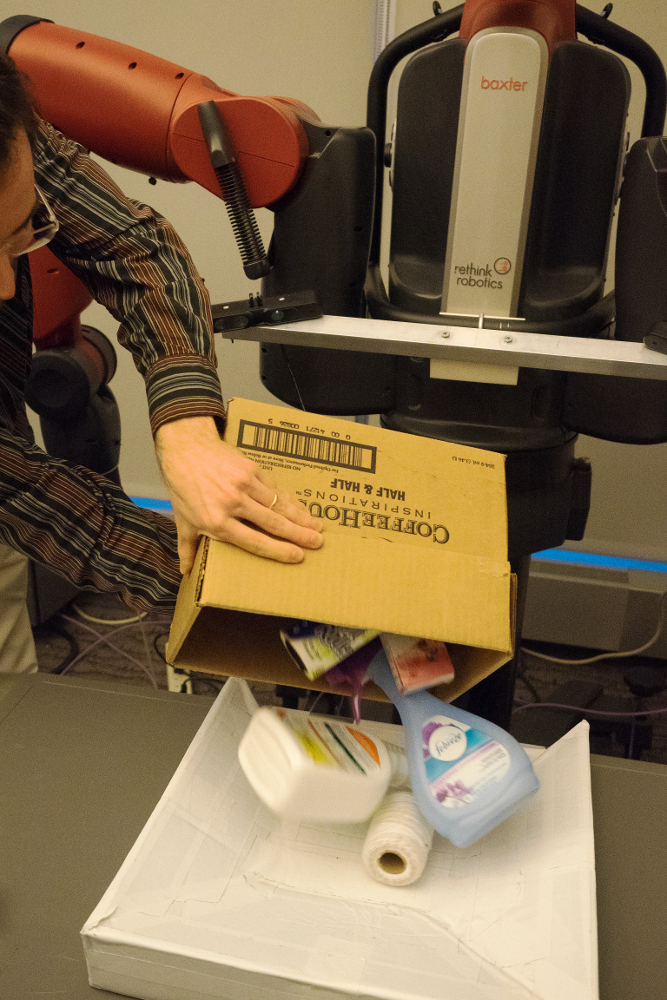}}
  \subfigure[]{\includegraphics[height=1.6in,trim={0.4in 0in 0.4in 0in},clip]{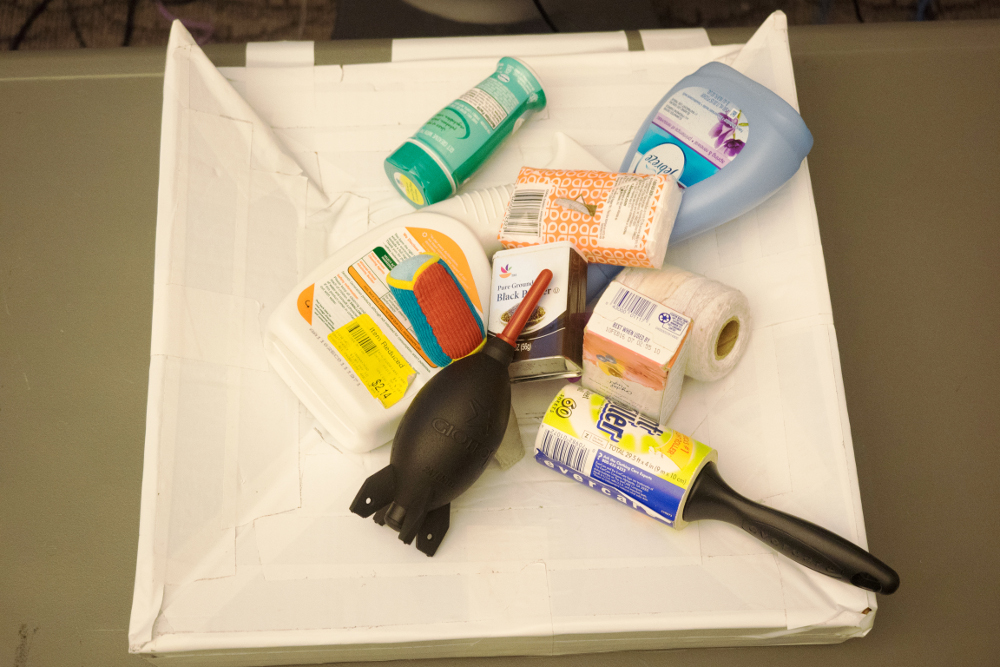}}
\end{center}
  \caption{Dense clutter benchmark task. (a) 10 objects are selected at random from a set of 27 total. (b) Pouring the box contents into the tray. (c) Tray contents immediately after pouring.}
  \label{fig:baxterDumpTray}
\end{figure}

The dense clutter benchmark task is illustrated in Figure~\ref{fig:baxterDumpTray} and the steps are outlined in Table~\ref{table:experimentProtocol}. First, ten objects are selected uniformly at random from the set of 27 shown in Figure~\ref{fig:baxterDumpTray}(a) (Step 1 in Table~\ref{table:experimentProtocol}). The 27 objects are common household items that are different from the 55 BigBird objects used for training. Next, the ten objects are placed in a box and the box is shaken to mix the objects (Steps 2 and 3 in Table~\ref{table:experimentProtocol}). Then, the contents of the box are poured into a tray placed in front of the robot on a table (Figure~\ref{fig:baxterDumpTray}(b), Step 4 in Table~\ref{table:experimentProtocol}). Then, the robot grasps as many objects one at a time as it can. The robotic grasping continues until either there are no objects remaining in the tray, the algorithm has run three times and no grasp hypotheses were found, or until the same failure occurs on the same object three times in a row. (The latter case only occurred once in our experiments, where the front of the vacuum attachment was not grasped in a stable way.)

\subsection{Hardware setup}

\begin{figure}[h]
\begin{center}  
  \subfigure[]{\includegraphics[height=1.8in]{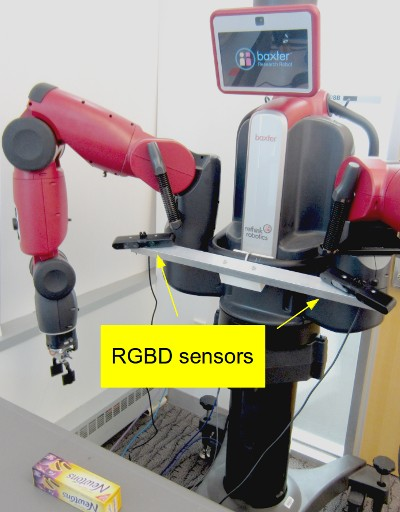}}
  \subfigure[]{\includegraphics[height=1.8in]{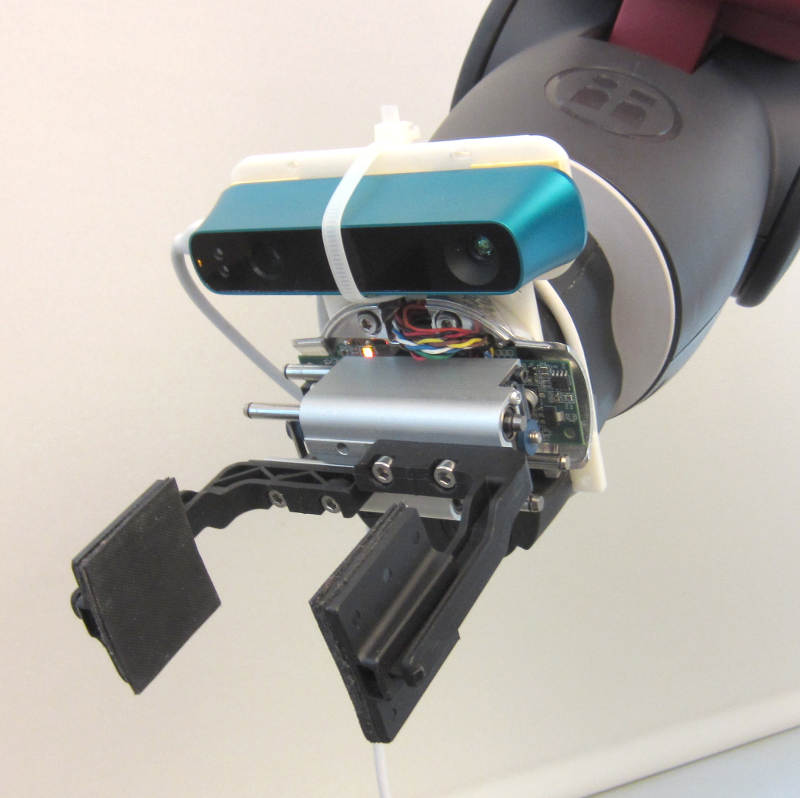}}
  \subfigure[]{\includegraphics[height=3.6in]{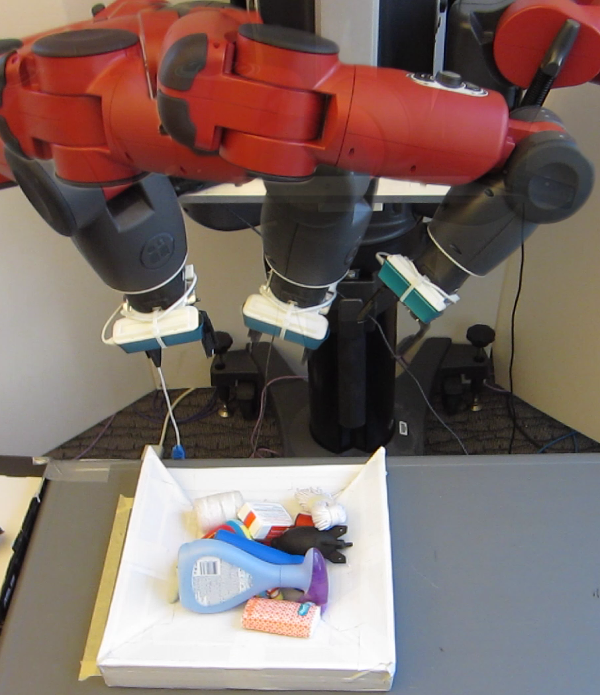}}
\end{center}
  \caption{(a) Baxter in a typical dense clutter test scenario. (b) We 
  mounted a depth sensor (Structure IO) to the robot hand to enable us 
  to obtain more complete point clouds using SLAM software. (c) Dense 
  clutter benchmark task scenario superimposed with an illustration of 
  three configurations along the trajectory taken by the arm during active sensing.}
  \label{fig:baxter_pic}
\end{figure}

\begin{figure}[h]
\begin{center}  
  \includegraphics[height=1.5in,trim={0in 0.2in 0in 0.4in},clip]{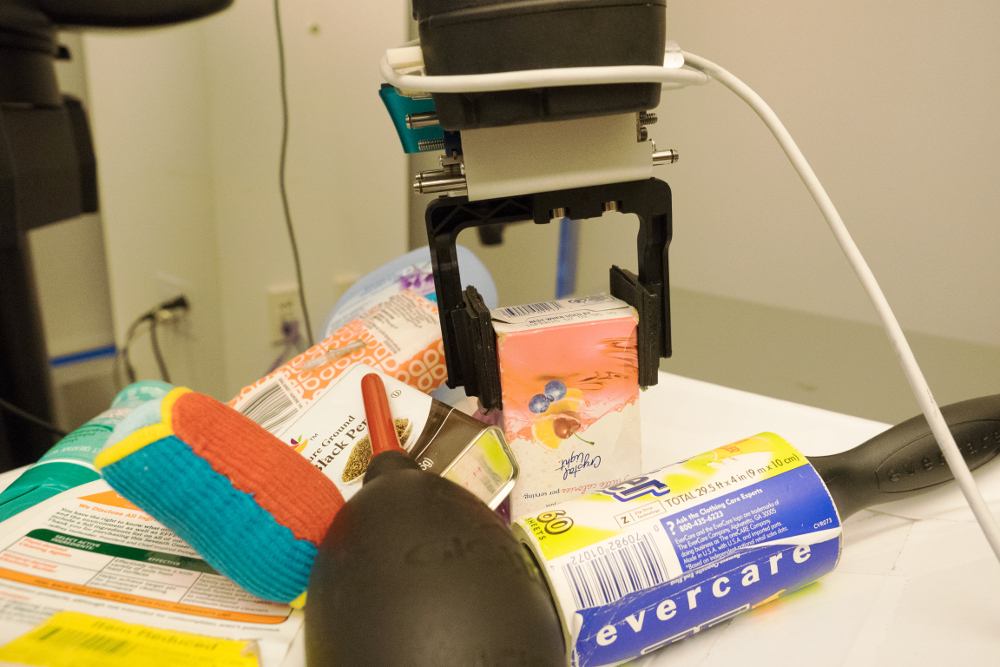}
\end{center}
  \caption{Gripper closing on the first object in the clutter.}
  \label{fig:baxterGrasp}
\end{figure}

We use the right 7-DOF arm of the Baxter Research Robot in the experiments (see Figure~\ref{fig:baxter_pic}). Our robot hand is the stock Baxter parallel-jaw gripper with the stock, short fingers and square pads. The square pads were modified with a black rubber covering, and rubber-covered pieces of metal were added to the ends (Figure~\ref{fig:baxterGrasp}). The ends bend slightly outward to initially widen the bite which helped with minor, sub-centimeter kinematic or point cloud registration errors. This gripper is restricted to a 3 to 7cm width. Each object in the test set was selected given this restriction. We mounted two Asus Xtion Pro depth sensors to Baxter's waist and an Occipital Structure IO sensor to the robot's wrist.

We used two computer systems in the experiments. Each system consisted of a 3.5 GHz Intel Corei7-4770K CPU (four physical cores), 32 GB of system memory, and an Nvidia GeForce GTX 660 graphics card. One system was used to run our GPD algorithm, and we used InfiniTAM~\citep{Kahler2015} on the other system to obtain a truncated signed distance function volume from the wrist-mounted sensor while moving the robot arm. Communication between the robot and the two PCs was handled by the robot operating system (ROS).

\subsection{Point Cloud Acquisition}

We explored two different strategies for acquiring a point cloud of the objects to be grasped: a passive strategy and an active strategy. The passive strategy used the two Asus Xtion Pro depth sensors mounted to the body of the robot as shown in Figure~\ref{fig:baxter_pic}(a). We measured the precise pose of each sensor in the base frame of the robot and used this information to register the two point clouds together. The active strategy creates a point cloud using off-the-shelf SLAM software (InfiniTAM~\citep{Kahler2015}). A depth sensor (Structure IO) is mounted near the end effector (Figure~\ref{fig:baxter_pic}(b)). Then, during point cloud acquisition, the arm moves through a collision-free trajectory on a hemisphere centered on the object pile while tracking using InfiniTAM. The trajectory is a 46cm geodesic between a fixed via point on one side of the pile and a fixed via point on the other. The radius of the hemisphere is 40cm (the minimum range of the Structure IO sensor is approximately 35cm). TrajOpt \citep{schulman2013finding} is used to generate the trajectory. At all times, the sensor is constrained to point directly towards a fixed point above the cluttered tray.

\subsection{Grasp Detection Runtime}

\begin{table}
\small\sf\centering
\caption{Run time averaged over 10 different scenes. \textit{Alg Version} denotes the representation number: 1 denotes the 15-channel representation; 3 denotes the 3-channel representation (see section entitled \textit{Comparison Between Different Representations}). \textit{Avg Num Pts} denotes the average number of points in the point cloud. \label{table:runtime}}
\begin{tabular}{|c||l|l|l|l|}
\hline
Alg Version & 1 & 1 & 3 & 3 \\
  \hline
Avg Num Pts & 39k & 66k & 39k & 66k \\
  \hline
  \hline
Avg time to sample & & & & \\
1k candidates & 0.8s & 1.7s & 0.8s & 1.7s \\
  \hline
Avg time to classify & & & & \\
1k candidates & 4.3s & 6.2s & 0.3s & 0.4s \\
\hline
\end{tabular}\\[10pt]
\label{table:runtime}
\end{table}

Since algorithm runtime is critical for real world applications, we have implemented our algorithm in C++ as a ROS package~\citep{gpd_website}. Both grasp candidate generation and classification are parallelized in our implementation. We implemented our neural network using Caffe~\citep{Jia2014}. We measured the runtime of the two key steps in our algorithm -- sampling and classification (lines 3 and 4-5 in Algorithm~\ref{alg:1}) for both the 15-channel
representation and the 3-channel representation. These runtime evaluations were run on an Intel Core i7-4770K Haswell Quad-Core 3.5GHz with 32GB of RAM and one NVIDIA GTX 970 GPU. We report the time required by the algorithm to do sampling and classification for 1k grasp candidates. Out of these 1k candidates, we typically find that the classifier is able to identify a few high-confidence grasps. However, times scale approximately linearly with candidates, so it is possible to detect additional grasps by increasing the time spent.

Table~\ref{table:runtime} shows results averaged over point clouds derived from ten different cluttered scenes similar to that shown in Figure~\ref{fig:baxterDumpTray}. There are four different contingencies. In the first column of Table~\ref{table:runtime}, we evaluate the runtime for the full 15-channel representation when run against point clouds produced by a single depth sensor containing on average 39k points. In the second column, we measure runtime for the 15-channel representation when run against point clouds produced by two registered depth sensors containing on average 66k points. In the third column, we measure runtime for the 3-channel representation when run against the same 39k point clouds as column 1. In the fourth column, we measure runtime for the 3-channel representation when run against the same 66k point clouds as column 2.

These results suggest two main conclusions. First, our proposed algorithm is practical for real world applications. In typical grasping scenarios, our algorithm can be expected to require between one and eight seconds of runtime, depending upon which version of the algorithm runs (columns two and three of Table~\ref{table:runtime}). Second, the 3-channel version of the algorithm runs much faster than the 15-channel version: the 3-channel version can handle the 66k point cloud in 2.1 seconds whereas the 15-channel version requires 8 seconds. As shown in Figure~\ref{fig:comparisonOfRepresentations}, going from the 15-channel representation to the 3-channel representation ``costs'' between 2 and 2.5 percentage points in accuracy. However, as shown above, the 3-channel version is four times faster. For this reason, we used the three-channel version of the algorithm in the robotic grasping experiments reported in the remainder of this section. We compensate for the lower accuracy by raising the grasp acceptance threshold (and thereby decreasing recall).

\subsection{Grasp Selection Strategy}
\label{sect:clutterRemovalAlgorithm}

In most grasp environments, grasp pose detection finds hundreds of grasps. Of these, we need to select one to execute. First, we prune the grasps that are infeasible. Since the Baxter gripper can only grasp surfaces that are between 3cm and 7cm thick (the Baxter gripper has a 4cm stroke), we prune grasps that are outside of the 3-7cm range. Second, we prune grasps that do not have collision-free inverse kinematics (IK) solutions. We use IKFast to generate the IK solutions and OpenRAVE for collision checking~\citep{Diankov2010} with obstacles created by centering 2cm cubes at points in the voxelized point cloud.

After pruning infeasible grasps as described above, several feasible grasps will typically still remain. Even though all of these are presumably all force closure grasps, they are not necessarily all equally good. In particular, task and object-related criteria will make some grasps more desirable than others. We handle this by creating heuristics based on our qualitative experience with the dense clutter domain. First, in the context of the dense clutter benchmark, there is a clear advantage to grasping objects at the top of the pile. Second, we have found top grasps to be more successful than side grasps. We encode both of these criteria by defining a cost function over the grasps. Third, since hand pose estimates based on forward kinematics for the Baxter arm are inaccurate, we penalize grasps that would require the arm to travel long distances in configuration space from a nominal configuration. This enables us to calibrate the arm for a smaller region of configuration space about the nominal configuration. These three criteria are expressed in the following cost function. Let $z(h)$ denote the height of hand $h$ along the gravity axis above the table. Let $\hat{a}(h)$ denote a unit vector pointing in the approach direction of the hand ({\em i.e.} along the negative $x$ axis of $F(h)$ as shown in Figure~\ref{fig:hand_coordinate_frame}). Let $\nabla q(h)$ denote the L2 (i.e. the Euclidean) distance in configuration space between the arm configuration at $h$ and some nominal starting configuration. The cost function we use for the dense clutter benchmark is:
\begin{equation}
\nonumber
J(h) = l\left(0.5(\hat{a}(h)^T \hat{g} + 1)\right) l\left(\frac{\nabla q(h)}{\nabla q_{m}}\right)
l\left(\frac{z_{m} - z(h)}{10z_{m}}\right),
\label{eqn:costfn}
\end{equation}
where $l(x) = \max(1-x,0)$ denotes a hinge loss, $z_{m}$ denotes the maximum height of a grasp above the table, $\hat{g}$ denotes a unit vector pointing in the direction of gravity, and $\nabla q_{m}$ the diameter. The grasp that scores the highest on this cost function is selected. The practical effect of these heuristics as implemented by Equation~\ref{eqn:costfn} is measured in the results presented in the next section (Table~\ref{table:results}). Without these heuristics (i.e. selecting grasps randomly instead), we get 18\% more grasp failures than otherwise. As discussed in the next section, these failures are caused primarily because of attempts at objects that are hard to reach or lift. Anecdotally, we have found that these heuristics are useful in any tabletop grasping scenario. They are easily adjusted to single-object grasping rather than clutter grasping simply by constraining the set of grasp candidates to lie only on an object of interest.


\subsection{Results}

\begin{table*}
\small\sf\centering
\caption{Results of the clutter-removal experiments.\label{table:results}}
\begin{tabular}{l|r|r|r|r}
\toprule
  & Active Point Cloud & Passive Point Cloud & No Selection Strategy & No Classification \\
  \hline
  Num objects & 300 & 150 & 150 & 150 \\
  \hline
  Num grasp attempts & 288 & 138 & 155 & 142 \\
  \hline
  Num grasp successes & 268 & 116 & 117 & 75 \\
  \hline 
  \% grasps successful & 93\% & 84\% & 75\% & 53\% \\
  \hline
  \% objects removed & 89\% & 77\% & 78\% & 50\% \\
\bottomrule
\end{tabular}\\[10pt]
\end{table*}

We evaluated performance against the dense clutter benchmark under 
four different contingencies (Table~\ref{table:results}): {\em 
active point cloud}, {\em passive point cloud}, {\em no selection 
strategy}, and {\em no classification}. In all cases, we used a CNN 
network trained on all 55 BigBird objects in our dataset. Grasp 
candidates were encoded using the 3-channel representation used in~
\citep{tenpas_isrr2015} (the red line in Figure~\ref
{fig:comparisonOfRepresentations}). Out of the 27 objects shown in 
Figure~\ref{fig:baxterDumpTray}(a), there were two objects that 
could not be grasped by the Baxter gripper in our configuration 
because of the 3cm min aperture when they fell into an upsidedown 
configuration: the green sandcastle and the red lobster. If either 
of these objects fell into this configuration during ``pouring'' of 
the objects into the tray (Step 4 of the dense clutter benchmark 
task as shown in Table~\ref {table:experimentProtocol}), then we 
manually removed the object, turned it right side up, and placed it 
back on the pile. 

In the {\em active point cloud} contingency, we obtained point clouds using 
the active sensing strategy where we run InfiniTAM while moving the 
wrist-mounted sensor above the tray. A point cloud with associated 
surface normals is obtained from the resulting truncated signed 
distance function. We used the grasp selection heuristics described 
above to select grasps. We ran 30 rounds of the dense clutter benchmark 
task where 10 objects were presented to the robot on each round. We 
observed a 93\% grasp success rate (20 grasp failures out of 288 
grasp attempts; the ``Active Point Cloud'' column of Table~\ref
{table:results})). Out of the 20 failures, 5 were due to point cloud 
registration errors or innaccuracies in the kinematic calibration of the robot, 9 were due to 
a failure of the algorithm to select what appeared to be a good grasp, 4 were due to a 
collision of the fingers with the object before the grasp, and 2 
were due to the object dropping out after an initially successful 
grasp. In this scenario, 89\% of the objects were cleared from the 
tray. 

In the {\em passive point cloud} evaluation contingency, we ran 15 
rounds of the dense clutter benchmark task where the point cloud was 
obtained using the passive strategy (two Asus Xtion Pro sensors 
fixed to the body of the robot). Otherwise, the algorithm was the 
same as in the {\em active point cloud} contingency. We observed an 
84\% grasp success rate (22 grasp failures out of 138 grasp 
attempts). Of the 22 failures, 5 were due to point cloud 
registration errors or innaccuracies in the kinematic calibration of the robot, 13 were due to 
a failure of the algorithm to select what appeared to be a good grasp, 2 were due to a 
collision of the fingers with the object before the grasp, and 2 
were due to the object dropping out after an initially successful 
grasp. In this scenario, 77\% of the objects placed in front of the 
robot were cleared. The others were either knocked out of the tray, 
pushed too far forward to be seen by the sensors, or grouped too 
close together for a finger-clear grasp to be found. We attribute 
the lower success rate (84\%) obtained here relative to that 
obtained using the active strategy (93\%) to two factors: 1) the 
point cloud created using the active strategy appears to better 
cover the objects in the box; 2) the point cloud created using the 
passive strategy does not see some objects near the very front of 
the box very well.

In the {\em no selection strategy} evaluation contingency, we ran 15 
rounds of the dense clutter benchmark task where we omitted the 
grasp selection strategy (Step 6 of Algorithm~\ref{alg:1}). 
Otherwise, the algorithm was the same as in the {\em active point 
cloud} contingency. We still pruned grasps that had no IK solutions, 
were in collision, or did not fit the Baxter gripper aperture 
constraints. However, instead of ranking the remaining grasps, we 
just selected grasps randomly. The second to last column of Table~\ref{table:results} shows the results. We observed a 75\% grasp success rate (38 grasp failures out of 155 grasp attempts). Out of the 38 failures, 9 were due to a failure of the algorithm to select what appeared to be a good grasp, 8 were due to collision of the fingers with the object prior to forming a grasp, 10 were due to point cloud 
registration errors or innaccuracies in the kinematic calibration of the robot, and 11 were due to the object 
dropping out after an initially successful grasp. Notice that, compared with the 
{\em active point cloud} contingency, there are a large number of 
grasps where the robot dropped the object after pickup, where there 
were errors caused by inaccurate kinematic calibration during the reach, or where there was a collision 
just prior to the grasp. This makes sense: since the heuristics help the 
algorithm avoid grasps at the bottom of the pile, they reduce these 
sorts of errors.

In the {\em no classification} contingency, we ran 15 rounds of the dense clutter benchmark task while omitting the grasp classification step (Step 5 of Algorithm~\ref{alg:1}). Instead of ranking and pruning low scoring grasps, we used the grasp selection strategy to select from among all sampled grasp candidates. This contingency implicitly assumes all sampled grasp candidates to be force closure grasps. Otherwise, the algorithm was the same as in the {\em active point cloud} contingency. The last column of Table~\ref{table:results} shows the results. We observed a 53\% grasp success rate (67 grasp failures out of 142 grasp attempts). Out of the 67 failures, 46 were due to a failure of the algorithm to select what appeared to be a good grasp, 6 were due to collision of the fingers with the object before the grasp, 14 were due to the object dropping out after an initially successful grasp, and 1 was caused by a point cloud registration error. These results reflect what one might expect. If we do not prune the grasp candidates that are not force closure grasps, then the algorithm suffers a large number of grasp failures caused by attempting to grasp poor candidates. In addition, this method produces many poor grasps that drop out of the hands during lifting.

\section{Combined Grasp and Object Detection}

\begin{figure*}
\begin{center}
  \subfigure[]{\includegraphics[height=1.5in]{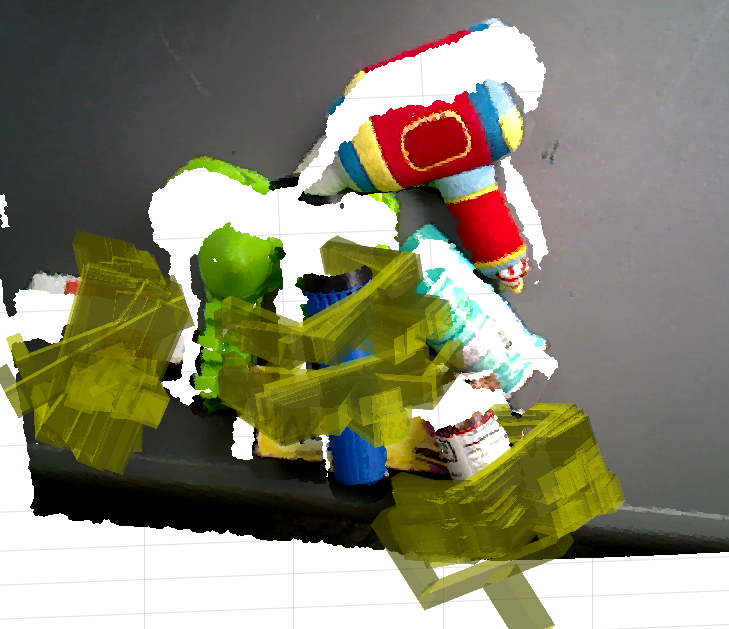}}
  \subfigure[]{\includegraphics[height=1.5in]{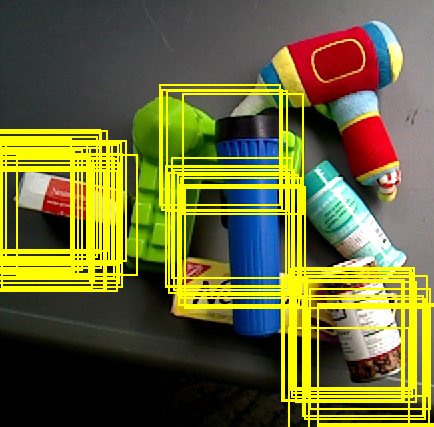}}
  \subfigure[]{\includegraphics[height=1.5in]{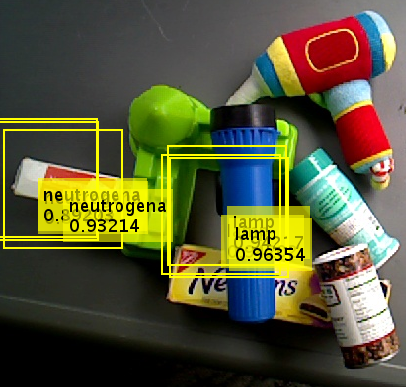}}
  \subfigure[]{\includegraphics[height=1.5in]{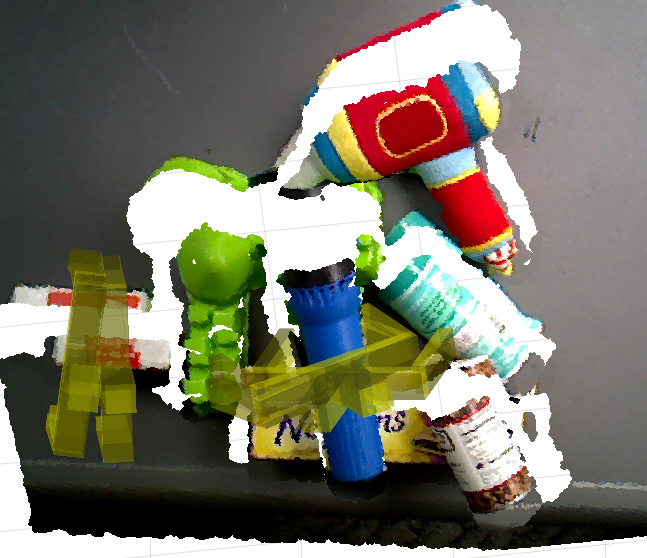}}
\end{center}
  \caption{Illustration of combined grasp and object detection. (a) a set of grasps detected in a scene; (b) the corresponding object proposals; (c) high-confidence object detections; (d) grasps corresponding to detected objects. For either of the two detected objects, the robot has the option to execute one of two detected grasps.}
  \label{fig:obj_detection}
\end{figure*}


\begin{figure}
\begin{center}  
  \includegraphics[height=1.8in]{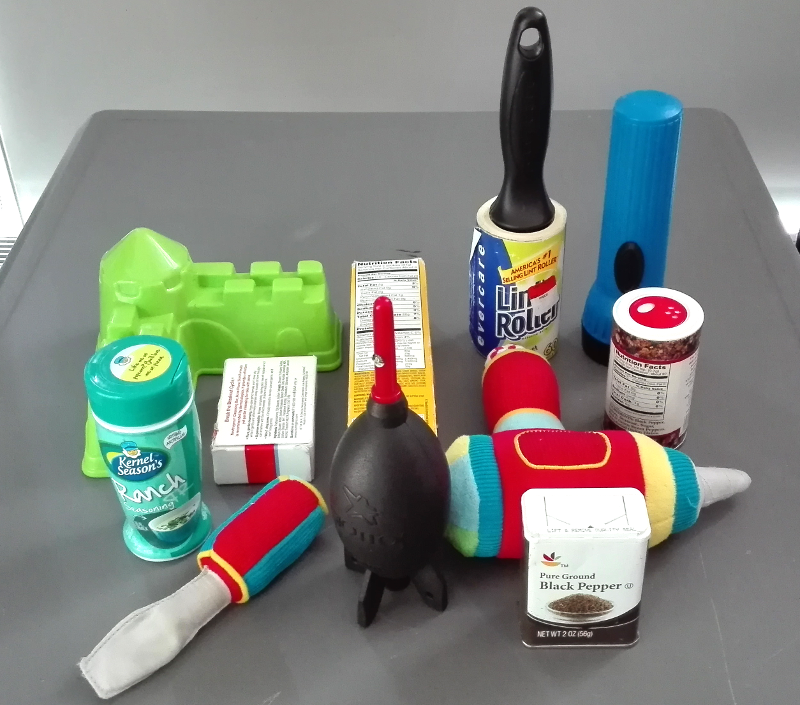}
\end{center}
  \caption{The 11 objects used to test combined object and grasp detection.}
  \label{fig:11obj}
\end{figure}

In a traditional grasp pipeline, e.g. where a CAD model of an object is registed to point cloud in 6DOF pose space, determining the identity of an object to be grasped is an integral part of the grasp perception process. However, grasp detection dispenses with this requirement. But, this does not mean that object identity is irrelevant. The opposite is nearly always true: we are almost always interested in grasping a particular object -- not just any object. This means that we must combine grasp detection with object detection. 

We propose a simple method of accomplishing this. The standard approach to object detection is to generate candidate object locations (i.e. object proposals) using methods like EdgeBoxes~\citep{zitnick2014edge} or selective search~\cite{uijlings2013selective} and then to classify these candidates using machine learning, i.e. a convolutional neural network. However, since we are primarily interested in classifying objects that can be grasped, we propose using the detected grasps themselves as object proposals. After detecting a set of potential grasps relative to the point cloud, we calculate a point directly between the fingers for each detected grasp and project this point into the RGB image. Then, we create the set of proposals by taking a fixed-size bounding box ($120 \times 120$ pixels in our case) around each projected point.

Figure~\ref{fig:obj_detection} illustrates the process. Figure~\ref{fig:obj_detection} (a) shows 68 grasps that were detected in the scene. Notice that grasps are only detected on three of the seven objects. There were no detections on the other four objects because they were beneath other objects or because they were too far away for the robot to reach. Figure~\ref{fig:obj_detection} (b) shows the corresponding object proposals in the RGB image. Notice that our use of the grasp detector as an object proposal generation mechanism focuses the attention of the system on graspable objects in a way that an off-the-self proposal strategy like EdgeBoxes or selective search would not. Figure~\ref{fig:obj_detection} (c) shows the top scoring object proposals in each object category that achieved a minimum threshold score. In this case, we obtained high confidence labels for the flashlight in the middle (the ``lamp'') and the white and red box on the left (``neutrogena''). Figure~\ref{fig:obj_detection} (d) shows the grasps corresponding to these object detections. Since we now have predictions of the object identity of all objects that can be grasped by the system, we can select an object to grasp as desired.

We performed experiments to evaluate the accuracy with which we can detect objects using this method. We trained the BVLC Reference Network (four convolutional layers and three fully connected layers)~\cite{Jia2014} to detect object instances for the set of 11 objects shown in Figure~\ref{fig:11obj}. In order to train the network, we automatically collected images for each of the 11 objects as follows. For each object, we first placed it alone on a table in front of the robot. Then, we collected a sequence of depth images from different perspectives by using a depth camera mounted to the end of our robotic arm. We repeated this process three times for each object, each time placing the object on the table in a different orientation. Altogether, we obtained approximately 2k images per object. We augmented this by rotating and scaling, resulting in a dataset of approximately 10k images per object -- approximately 116k images total. We created a train/test set with a 75\%/25\% split. Starting with a version of the BVLC Reference Network pretrained on ImageNet, we finetuned the network on our 11 objects by running 20k iterations of stochastic gradient descent where we used a learning rate of 0.000001, momentum of 0.9, weight decay of 0.0005, and where each minibatch contained 128 images. After training, we obtained more than 99\% accuracy on the test set. This 99\% accuracy can be interpreted as the expected performance of the object classifier in a single-object grasp setting.

\begin{table}
\small\sf\centering
\caption{Confusion matrix for the nine objects used to test combined grasp and object detection. Object/letter correspondences shown below.\label{table:confusion}}
\begin{tabular}{l|lllllllll}
     &A&B&C&D&E&F&G&H&I    \\
     \hline
     A&17   &  0  &   0 &    0  &   0  &   0  &   0  &   0  &   0    \\
     B&0   & 18  &   0 &    0  &   5  &   0  &   1  &   0  &   0   \\
     C&0   &  0  & 136 &    0  &  14  &   0  &   0  &   0  &   0   \\
     D&1   &  0  &   0 &  228  &   0  &   0  &   0  &   0  &   0   \\
     E&0   &  0  &   0 &    0  & 206  &   0  &   3  &   0  &  15   \\
     F&8   &  4  &  28 &    0  &  35  & 150  &   3  &   0  &  16  \\
     G&0   &  0  &   0 &    0  &   0  &   0  &  76  &   0  &   0   \\
     H&0   &  0  &   8 &    0  &  31  &   1  &   0  & 136  &   1   \\
     I&0   &  0  &   0 &    0  &   0  &   0  &   0  &   0  &  56   \\
\end{tabular}\\[10pt]

\begin{tabular}{l|l}
     Letter & Label \\
     \hline
     A& Green Castle \\
     B& Plush Drill \\
     C& Fig Newtons Box \\
     D& Blue Flashlight \\
     E& Lintroller \\
     F& Neutrogena Box \\
     G& Cyan Ranch Dressing Bottle \\
     H& Cylindrical Red Pepper Bottle \\
     I& Rocket \\
\end{tabular}
\end{table}

We wanted to evaluate the performance of our approach in cluttered scenarios. To accomplish this, we created a set of 9 cluttered scenes involving 9 of the 11 objects (similar to that shown in Figure~\ref{fig:obj_detection}) -- we excluded the screwdriver because it was hard to grasp using the Baxter hand and the rocket because it was nearly indistinguishable from the link roller handle for the $120 \times 120$ window size used in this work. We took 3 point clouds of each of the 9 cluttered scenes, for a total of 27 point clouds. For each point cloud, we used grasp detection to create a set of object proposals, for a total of 1197 object proposals. We manually labeled each proposal with the identity of the object that would get picked up if the robot were to execute the corresponding grasp. Then, we evaluated the accuracy of the object classifier over this set of object proposals. Out of the 1197 total grasps, our system correctly predicted the correct object in 1023 of these (85.4\% accuracy). Table~\ref{table:confusion} shows the corresponding confusion matrix. The primary reason why object classification accuracy drops in clutterrelative to the single object scenario is that the object proposals sometimes contain adjacent objects in addition to the objects that would be grasped. By using segmentation to remove these adjacent objects from the proposals, we expect that it should be possible to improve on the results above. Nevertheless, the fact that we get 85\% average object detection accuracy in relatively cluttered settings suggests that the approach above should be practical in some application settings as is.

An important application of object detection is to detect object class in order to improve grasp detection accuracy. Recall that we showed earlier that it is possible to improve grasp detection accuracy by training a neural network using only grasps from the specific object instance or object category of a target object. In this context, the object detection method described in this setion could be used to predict object instance or category as part of this process. Given a novel scene, we might generate grasp candidates for all visible objects and then predict object instance or category for each grasp. Then, given this prior, we would invoke the appropriate classifier to detect grasps on the object. This approach would be particularly useful in scenarios where it is desired to grasp an object based on its category label. One would eliminate grasp candidates that were not predicted to have the desired class label and focus classification on just the remainder.

\section{Discussion}

This paper makes a case for grasp detection as an alternative to attempting to estimate the exact pose of objects to grasp. This is exactly what many approaches to grasp perception try to do. For example, the ROS grasp pipeline registers a geometric model of the object to be grasped to a point cloud \citep{chitta_ram2012}. Unfortunately, registering a CAD model to a noisy or incomplete point cloud can be very challenging. Moreover, the assumption that a CAD model will always be available for every object that a robot might need to grasp is incompatible with our desire to operate in open world environments. Instead, grasp detection finds grasp poses directly in a point cloud or truncated signed distance function. In contrast to the object-pose-estimation-approach, grasp detection: 1) does not require any object segmentation; 2) does not need to know the identity, class, or shape of the object to be grasped; 3) does not try to register a CAD model of the object to the point cloud. This is a significant advantage: object segmentation and 3D registration are both difficult problems, and a failure in either will cause traditional grasp perception to fail. We attribute the high grasp success rate that we achieve on the dense clutter benchmark to these advantages. The fact that these grasp success rates are averages over a large number of grasp trials (all together, our grasp results are averages over 723 dense clutter grasp experiments with 288 of these grasp attempts performed in our best-case contingency) suggest that these success rates are repeatable (at least for the objects we used in our experiments, shown in Figure~\ref{fig:baxterDumpTray} (a)). 

We performed several experiments in simulation and on the hardware to quantify the performance contributions of various parts of the system. The results suggest that several different pieces are needed in order for grasp detection to work well. Representation of the grasp candidate is important: our results (Figure~\ref{fig:comparisonOfRepresentations}) show that the ``right'' representation can improve classification accuracy by as much as 10\%. In addition, pretraining on simulated point clouds and using prior knowledge about the object to be grasped can help. Our experiments with the robotic hardware (Table~\ref{table:results}) show that improving the quality of the point cloud can also make a big difference. Grasp success rates where the point cloud was created using ``active sensing'' ({\em i.e.} using off-the-shelf metric SLAM software) are 9\% higher than when grasping using a point cloud obtained from sensors fixed to the robot body. In addition, grasp selection strategy is extremely important. As the third column of Table~\ref{table:results} shows, applying simple heuristics to choose which grasp to try first ({\em i.e.} grasp object at the top of the pile first) can improve grasp success rates by nearly 20\%. Finally, as the last column of Table~\ref{table:results} shows, grasp classification and ranking itself is very important. Without a mechanism for predicting which grasps are force closure, grasp success rates can drop by as much as 40\%.

\subsection{Limitations}

\begin{figure}
\begin{center}  
  \includegraphics[height=1.5in]{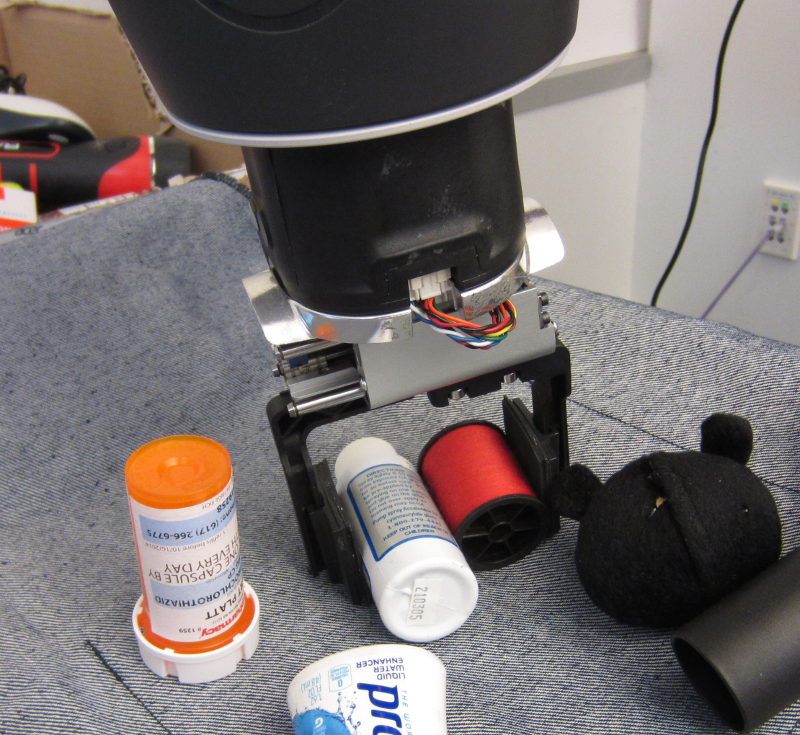}
\end{center}
  \caption{One grasp detection failure mode: two objects are grasped as if they were a single object.}
  \label{fig:doublegrasp}
\end{figure}

There are at least two drawbacks to grasp detection relative to more standard approaches. The first is illustrated in Figure~\ref{fig:doublegrasp}. Since our algorithm does not segment the objects, it can detect ``grasps'' that treat multiple objects as a single atomic object. This type of error is unusual with small-aperture hands, but one would expect it to become a more significant problem when the hand is physically capable of grasping larger objects. One approach to solving this problem would be to use object segmentation.

Another concern is whether grasp detection methods can be used to grasp specific objects of interest reliably. As we have shown in the last section, grasp detection can be combined with object detection, resulting in a system that can grasp specific objects of interest. However, it is not yet clear how accurately this can be done in general. Our experiments showed that this approach can work with 85\% object classification accuracy in a laboratory setting. It may be possible to improve this by incorporating segmentation methods. Regardless, improving upon this 85\% object classification accuracy result will be critical for deploying this type of system in real world settings.

\begin{acks}
The authors gratefully acknowledge support from the NSF (IIS-1427081), NASA (NNX13AQ85G and NNX16AC48A), and ONR (N00014-14-1-0047). We also thank the anonymous reviewers for great comments.
\end{acks}

\bibliographystyle{./SageH.bst}
\bibliography{platt}


\end{document}